\documentclass{article}

\usepackage{arxiv}

\usepackage[utf8]{inputenc}
\usepackage{amsmath,amssymb}
\usepackage[hidelinks]{hyperref}%
\usepackage{url}
\usepackage{doi}%
\usepackage{graphics,graphicx}
\usepackage{amsmath,amssymb}
\usepackage{geometry}
\usepackage{bm}
\usepackage{tikz}
\usepackage{subcaption}
\usepackage{comment}
\usepackage{cuted}
\usepackage{booktabs}
\usepackage{multirow}
\usepackage{lscape}
\usepackage{longtable}
\usepackage{mathtools}
\usepackage{cite}

\newenvironment{todo}{\color{red}\begin{itshape}}{\end{itshape}\color{black}}

\title{Towards Understanding the Survival of Patients with High-Grade Gastroenteropancreatic Neuroendocrine Neoplasms: An Investigation of Ensemble Feature Selection in the Prediction of Overall Survival}
\renewcommand{\shorttitle}{Feature Selection in High-Grade Gastroenteropancreatic Neuroendocrine Neoplasms}

\author{
\href{https://orcid.org/0000-0002-6919-3483}{\includegraphics[scale=0.06]{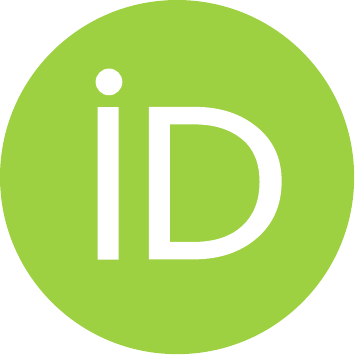}\hspace{1mm}
Anna Jenul}\\
Department of Data Science\\
Norwegian University of Life Sciences\\
Ås, Norway\\
\texttt{anna.jenul@nmbu.no}\\
\And
\href{https://orcid.org/0000-0002-2861-0340}{\includegraphics[scale=0.06]{orcid.pdf}\hspace{1mm}
Henning Langen Stokmo}\thanks{Henning Langen Stokmo is also affiliated with the Institute of Clinical Medicine, University of Oslo, Oslo, Norway}\\
Division of Radiology and Nuclear Medicine\\
Oslo University Hospital \\
Oslo, Norway\\
\texttt{h.l.stokmo@studmed.uio.no}\\
\And
\href{https://orcid.org/0000-0003-1327-4855}{\includegraphics[scale=0.06]{orcid.pdf}\hspace{1mm}
Stefan Schrunner}\\
Department of Data Science\\
Norwegian University of Life Sciences\\
Ås, Norway\\
\texttt{stefan.schrunner@nmbu.no}
\And
\href{https://orcid.org/0000-0003-3300-7420}{\includegraphics[scale=0.06]{orcid.pdf}\hspace{1mm}
Mona-Elisabeth Revheim}\thanks{Mona-Elisabeth Revheim is also affiliated with The Intervention Centre, Oslo University Hospital, Oslo, Norway, and the Institute of Clinical Medicine, University of Oslo, Oslo, Norway
}\\
Division of Radiology and Nuclear Medicine\\
Oslo University Hospital \\
Oslo, Norway\\
\texttt{monar@ous-hf.no}\\
\And
\href{https://orcid.org/0000-0003-3454-7461}{\includegraphics[scale=0.06]{orcid.pdf}\hspace{1mm}
Geir Olav Hjortland}\\
Department of Oncology\\
Oslo University Hospital\\
Oslo, Norway\\
\texttt{goo@ous-hf.no }\\
\And
\href{https://orcid.org/0000-0003-1595-9962}{\includegraphics[scale=0.06]{orcid.pdf}\hspace{1mm}
Oliver Tomic}\\
Department of Data Science\\
Norwegian University of Life Sciences\\
Ås, Norway\\
\texttt{oliver.tomic@nmbu.no}}

\begin{document}

\maketitle

\begin{abstract}
    Determining the most informative features for predicting the overall survival of patients diagnosed with high-grade gastroenteropancreatic neuroendocrine neoplasms is crucial to improve individual treatment plans for patients, as well as the biological understanding of the disease. Recently developed ensemble feature selectors like the Repeated Elastic Net Technique for Feature Selection (RENT) and the User-Guided Bayesian Framework for Feature Selection (UBayFS) allow the user to identify such features in datasets with low sample sizes. While RENT is purely data-driven, UBayFS is capable of integrating expert knowledge a priori in the feature selection process. In this work we compare both feature selectors on a dataset comprising of 63 patients and 134 features from multiple sources, including basic patient characteristics, baseline blood values, tumor histology, imaging, and treatment information. Our experiments involve data-driven and expert-driven setups, as well as combinations of both. We use findings from clinical literature as a source of expert knowledge. Our results demonstrate that both feature selectors allow accurate predictions, and that expert knowledge has a stabilizing effect on the feature set, while the impact on predictive performance is limited. The features \textit{WHO Performance Status}, \textit{Albumin}, \textit{Platelets}, \textit{Ki-67}, \textit{Tumor Morphology}, \textit{Total MTV}, \textit{Total TLG}, and \textit{SUV$_{\max}$} are the most stable and predictive features in our study.   
\end{abstract}

\section{Introduction}

Gastroenteropancreatic (GEP) neuroendocrine neoplasms (NEN) are heterogeneous types of malignancies increasingly common over the last three decades~\cite{henning01,henning02}. High-grade GEP NEN encompasses both neuroendocrine tumors grade 3 (NET G3) and neuroendocrine carcinomas (NEC), where NEC is further subdivided into small cell (SC) and large cell carcinomas (LC). According to the WHO 2019 Classification of Tumors: Digestive System Tumors, NET G3 are well differentiated (WD), whilst NEC are poorly differentiated (PD), both with a Ki-67 proliferation index (Ki-67) $>20\%$~\cite{henning03}. Although both NET G3 and NEC share features of immunohistochemical staining with chromogranin A and synaptophysin, they are considered morphologically different~\cite{henning04}.

The prognosis for patients with advanced GEP NEC is poor, with a median survival of less than 12 months~\cite{henning05,henning06}, whilst the prognosis for locoregional GEP NEC is higher; 20.7 months~\cite{henning07}. Numerous recently published studies~\cite{henning05,stokmo2022gastro,henning09,henning10,henning11,henning12,henning13,henning14} have shown the prognostic importance of several parameters on overall survival (OS) such as age, performance status (PS), primary tumor site, tumor differentiation, TNM-stage, serum lactate dehydrogenase (LDH), serum platelet levels, proliferation marker Ki-67, maximum standardized uptake value (SUV$_{\max}$), total metabolic tumor volume (tMTV) and total total lesion glycolysis (tTLG). Establishing more robust prognostic parameters and validating established parameters is essential to provide optimal care for this patient group.

Forecasting the OS of cancer patients as a major indicator of treatment success by machine learning models is of high relevance to offering optimal individual treatments for patients. In particular, accurate outcome prediction models pave the way for decision support in clinical practice. Since GEP NEN are rare, however, the data basis for training purely data-driven models is limited, leading to problems like overfitting, spurious correlations, and, consequently, to inaccurate predictions~\cite{kubben2019fundamentals,welch2019vulnerabilities,wallis2022brain}. Two major approaches are at hand to overcome these issues: (a) increasing the number of samples (either by collecting more data or by artificial data augmentation) or (b) reducing the dimensionality of the feature space. In this work, we elaborate on approach (b), where our method of choice is feature selection. While general dimensionality reduction methods like Principal Component Analysis~\cite{jolliffe2016pca} transform the data to a new domain and thereby make identification of influencing factors difficult, feature selection reduces the dimension by subsetting the dataset by columns. As a result, a subset of the original features is retained, and the interpretability of the data columns is preserved.

Beyond the obvious benefit that predictive models become tractable, feature selection has the potential to improve the understanding of biological processes by clinical experts~\cite{jenul2022ubayfs}. In particular, feature selectors point to input data parameters, which are related to explaining the target variable by a data-driven model. This information may either support or contradict existing hypotheses about the underlying biological processes or disclose previously unknown relations. The evaluation and interpretation of the findings require close collaboration between clinical experts and data scientists. However, such an application of feature selectors is still less common in machine learning, where the focus typically lies exclusively on optimizing performance metrics.

State-of-the-art research in feature selection with applications in healthcare, such as L1 regularization~\cite{tibshirani1996lasso}, decision trees~\cite{breiman1984decisiontrees}, Laplace scores~\cite{he2005laplacian}, or the minimum redundancy-maximum relevance (mRMR) criterion~\cite{ding2005mrmr}, are mainly data-driven and may suffer from well-known limitations. Among these limitations is the problem that minor changes, such as the inclusion of new or removal of old samples, may have significant effects on the set of selected features --- the property of feature sets to remain invariant under such changes to the dataset is referred to as feature selection stability and investigated in \cite{nogueira2018stability}. The usage of ensemble feature selectors, which train multiple feature selectors on subsets of the samples in a dataset, has recently been investigated extensively~\cite{bolon2018ensemble} and achieves a higher feature selection stability compared to a single feature selection run, while retaining a similar predictive performance, as used e.g., in random forest methods~\cite{breiman2001randomforest}. More recently, this fact has been exploited to introduce more stable feature selection methods tailored for healthcare applications, which offer a large potential with respect to the aspects discussed above~\cite{jenul2021rent,jenul2022ubayfs}.


This paper aims to improve the understanding and insights into the OS in patients with high-grade GEP NEN by applying recently developed ensemble feature selection techniques. Specifically, we evaluate the Repeated Elastic Net Technique for Feature Selection (RENT)~\cite{jenul2021rent}, as well as the User-Guided Bayesian Framework for Feature Selection (UBayFS)~\cite{jenul2022ubayfs} on a dataset containing 63 patients diagnosed with high-grade GEP NEN. Our experiments compare both ensemble feature selectors in setups with and without the use of expert information. Our main goals are: (I) to determine the most informative set of features with respect to the outcome prediction task; (II) to interpret those selected features clinically --- to evaluate the first goal, we measure the quality of the selected feature set in terms of predictive performance and selection stability. Another aspect of interest is: (III) to determine the effect of integrating prior expert knowledge into the feature selection process, compared to a purely data-driven pipeline. To this end, we discuss the feature selection results with respect to their clinical relevance and potential to improve our understanding of what influences OS of GEP NEN patients.

\paragraph{Notations} In the following, we denote the input data matrix by $\bm{X}\in \mathbb{R}^{m\times n}$, where $m$ denotes the number of patients, and $n$ denotes the number of features. Further, the target variable is denoted by $\bm{y}\in \mathbb{R}^m$. A feature set $S$ is characterized by the indices, $S\subseteq \{1,\dots,n\}$. Vectors and matrices are indicated by bold letters.

\section{Materials and Methods}

\subsection{GEP NEN dataset}

\paragraph {Statements of Ethics} This study was done in concordance with the Declaration of Helsinki. Approval from the regional committee for medical and health research ethics (2012/490, 2012/940, 2018/1940) and the local data protection officer was obtained. Informed consent was obtained from all patients at the time of inclusion but was waived for the patients in terminal phase and deceased.

\paragraph {Patient cohort} Patients were identified from a single institutional cohort at Oslo University Hospital, also included in two multi-institutional Nordic NEC registries organized by the Nordic Neuroendocrine Tumor Group, previously described by \cite{stokmo2022gastro}. In short, this cohort consisted of 192 patients included between January 2000 and July 2018, with GEP NEC classified according to the WHO 2010-classification~\cite{henning15}. In addition, all patients who had performed a fluorine-18 labeled 2-deoxy-2-fluoroglucose ([18F]FDG) positron emission tomography/computed tomography (PET/CT) within 90 days of their histological evaluation were eligible for inclusion. A hundred and seven patients did not have PET/CT performed, and two patients had no metabolic active lesions available for evaluation. Seventeen patients had more than 90 days between their biopsy and PET/CT, leaving 66 patients available for inclusion in this study.

\paragraph {Histological re-evaluation} As described previously in \cite{stokmo2022gastro}, the histological re-evaluation was performed on both core biopsies and surgical specimens from GEP NEC primary tumors and metastases. These were re-classified according to the most recent WHO 2019-classification~\cite{henning03} with regards to synaptophysin, chromogranin A, and the proliferation marker Ki-67. In this study, only the re-evaluated histology features were used, while the original histology block was discarded.

\paragraph {PET/CT acquisition} All PET/CT scans were done according to the European Association of Nuclear Medicine (EANM) guidelines \cite{henning04,henning05} as part of the clinical routine. The three PET scanners used were a 40-slice Siemens Biograph mCT hybrid PET/CT system (Siemens Healthineers, Erlangen, Germany), a Siemens Biograph 64, and a 64-slice General Electric (GE) Discovery 690 (GE Healthcare, Waukesha, WI, USA). Both Biograph PET/CTs were both EANM Research Ltd. (EARL)-accredited, whilst the Discovery 690 followed similar routine quality controls harmonizing with the two Biographs for cross-calibration. All acquisitions were from the vertex or skull base to mid-thighs. Before the PET acquisition, a low dose CT was acquired for anatomical information and attenuation correction. Parameters from PET were extracted using the ROI Visualisation, Evaluation, and Image Registration (ROVER) software v3.0.5 (ABX GmbH).\footnote{The detailed imaging- and extraction protocol is described in \cite{stokmo2022gastro}.}

\paragraph {Treatment} All patients received treatment in the form of surgery, chemotherapy, or a combination of both. In total, 54 patients received the standard treatment of platinum-based chemotherapy. Patients could have surgery prior to or after [18F]FDG PET/CT. Evaluation of response to chemotherapy treatment was done with CT using the Response Evaluation Criteria in Solid Tumors (RECIST)~\cite{henning16}.

\begin{figure}
    \centering
    \includegraphics[scale=0.5]{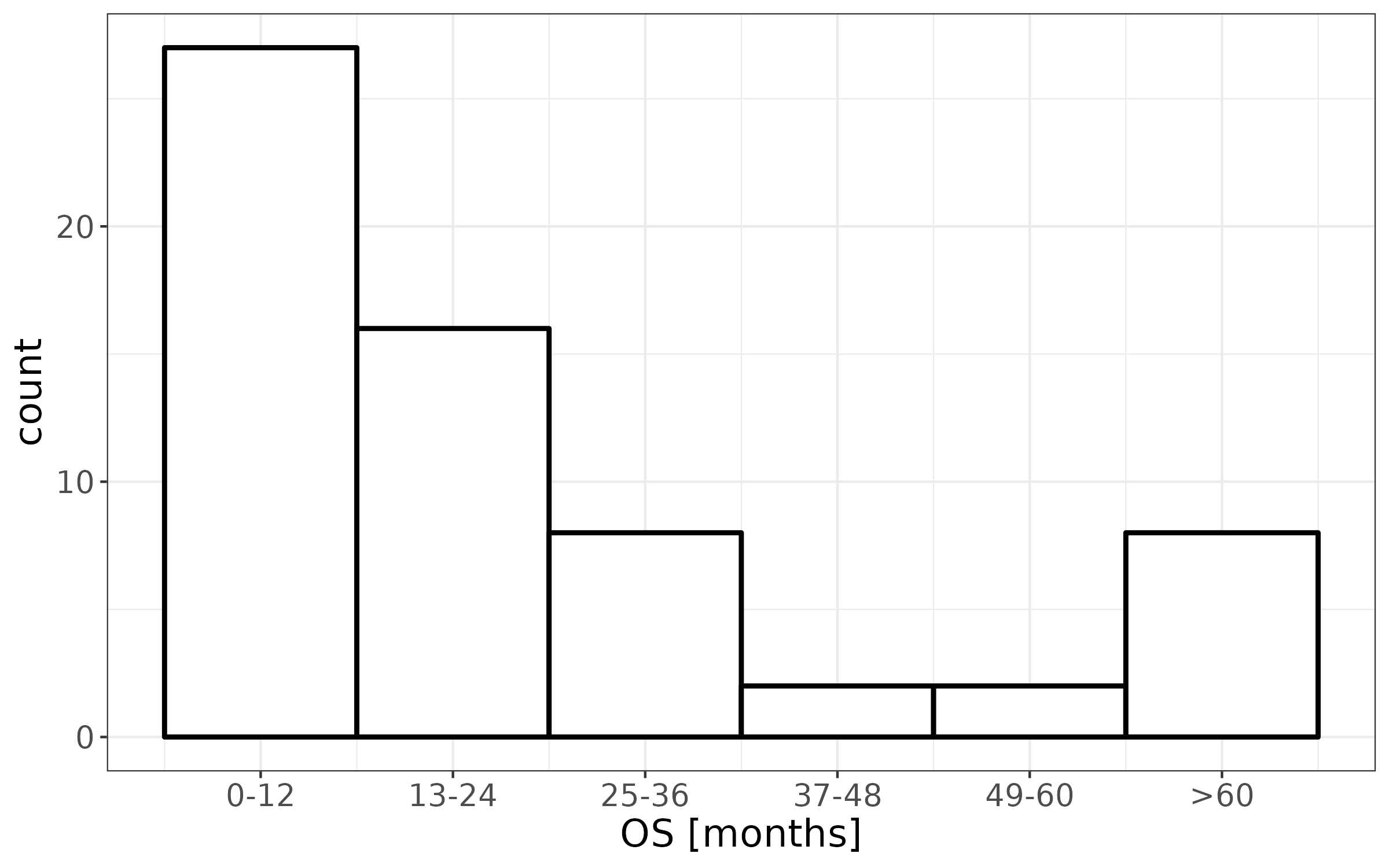}
    \caption{Distribution of the overall survival in months.}
    \label{fig:target}
\end{figure}

\paragraph {Outcome variable} Our outcome variable, or outcome target, was overall survival (OS) in months. This can be defined as the time a patient remains alive from the time of diagnosis to death of any cause; hence, it is not disease-specific. It is a reliable and easily available survival measure~\cite{henning17}. We can analyze such survival data, i.e., the time from diagnosis to the time of death, using the Kaplan-Meier estimator. For those patients who did not experience the event during the time of the study (or during follow-up) (i.e., death), they are said to be 'censored'~\cite{henning18}. Being 'censored' means that we do not know when this event will occur, only that it has not happened at the end of the study (or during follow-up). Across the full dataset, the empirical distribution of the outcome variable is illustrated as a histogram in Fig.~\ref{fig:target}.

\begin{figure}
    \centering
    \includegraphics[width=\textwidth, trim = 18 60 12 50, clip]{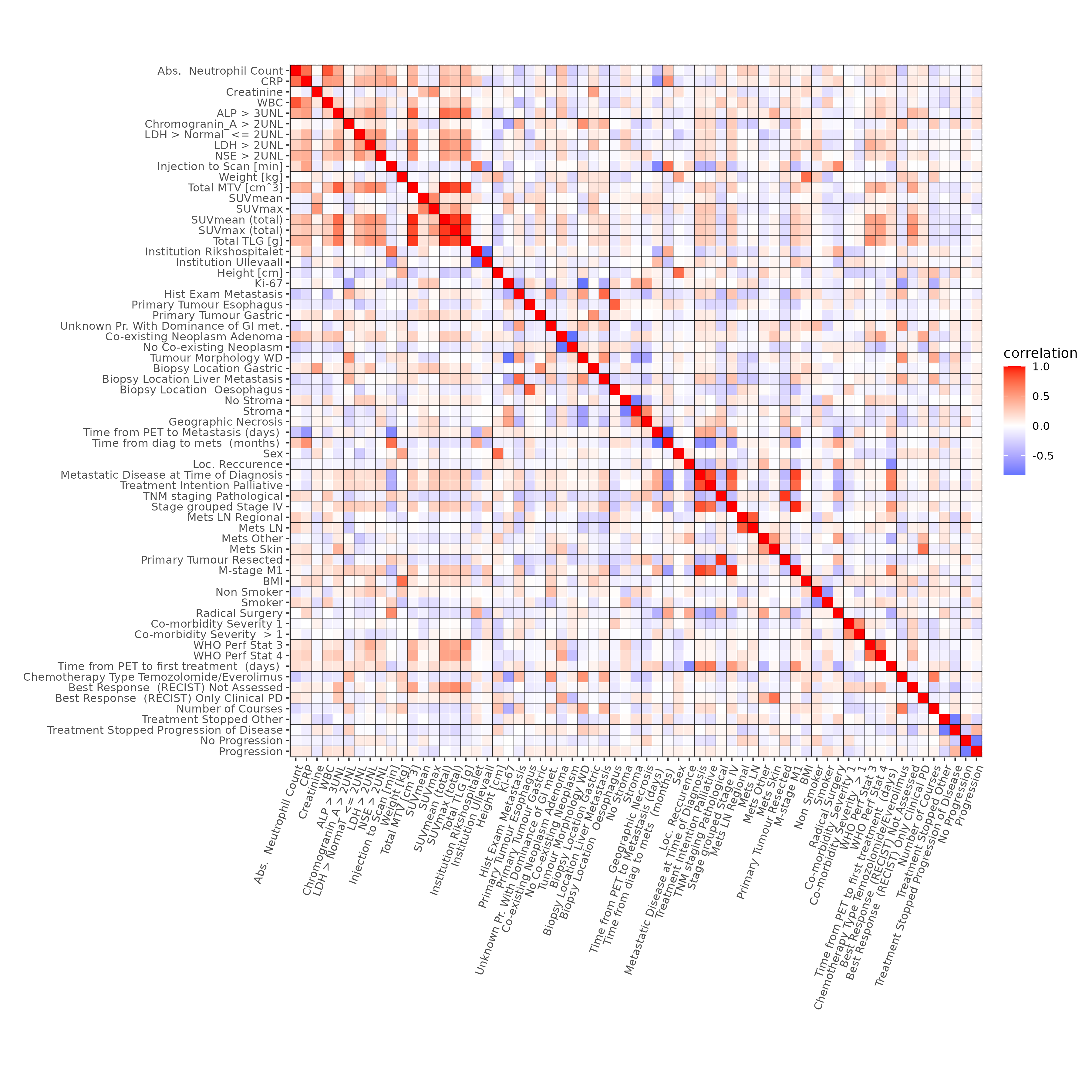}
    \caption{Correlations between input features (features with absolute correlations $\leq0.5$ were removed).}
    \label{fig:correlations}
\end{figure}

\newpage
\paragraph {Data blocks} The data were grouped into five different blocks
\begin{itemize}
    \item[(p)] patient characteristics
    \item[(b)] baseline blood values
    \item[(h)] re-evaluated histology
    \item[(i)] PET/CT imaging
    \item[(t)] treatment
\end{itemize}
The data contained mainly categorical and ordinal features with very few continuous variables. An overview of the pairwise correlations between the $n=134$ features is provided in Fig.~\ref{fig:correlations}.

\subsection{Data preprocessing}
\label{sec:preprocessing}

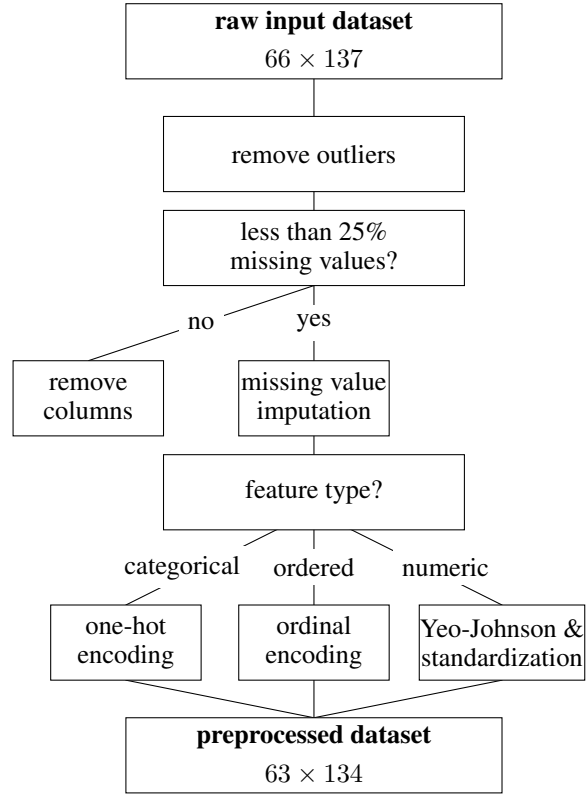
\begin{figure}
    \centering
     \begin{tikzpicture}
        \draw (0,0) -- (0,-2.75) -- (-3,-3.75);
        \draw (0,-5.75) -- (-2.5,-7) -- (-2.5,-8) -- (0,-8.5);
        \draw (0,-2.75) -- (0,-3.75) -- (0,-4.75) -- (0,-5) -- (0,-8.5);
        \draw (0,-5.75) -- (2.5, -7) -- (2.5,-8) -- (0, -8.5);
        \node[fill=white,align=center] at (0,-3.25) {yes};
        \node[fill=white,align=center] at (-1.5,-3.25) {no};
        \node[fill=white,align=center] at (-1.75,-6.5) {categorical};
        \node[fill=white,align=center] at (0,-6.5) {ordered};
        \node[fill=white,align=center] at (1.75,-6.5) {numeric};
        
        \draw (-2.5,0) rectangle ++(5,1);
        \node at (0,0.75) {\textbf{raw input dataset}};
        \node at (0,0.25) {$66 \times 137$};

        \draw[fill=white] (-2,-1.5) rectangle ++(4,1);
        \node at (0,-1) {remove outliers};
        \draw[fill=white] (-2,-2.75) rectangle ++(4,1);     
        \node[align=center] at (0,-2.25) {less than 25\% \\ missing values?};
        \draw[fill=white] (-4,-4.75) rectangle ++(2,1);      
        \node[align=center] at (-3,-4.25) {remove \\ columns}; 
        \draw[fill=white] (-1,-4.75) rectangle ++(2,1);      
        \node[align=center] at (0,-4.25) {missing value \\ imputation};
        
        \draw[fill=white] (-2,-6) rectangle ++(4,1);
        \node at (0,-5.5) {feature type?};
        \draw[fill=white] (-3.5,-8) rectangle ++(2,1);      
        \node[align=center] at (-2.5,-7.5) {one-hot \\ encoding};
        \draw[fill=white] (-1,-8) rectangle ++(2,1);      
        \node[align=center] at (0,-7.5) {ordinal \\ encoding};
        \draw[fill=white] (1.4,-8) rectangle ++(2.2,1);      
        \node[align=center] at (2.5,-7.5) {Yeo-Johnson \& \\ standardization};  

        \draw[fill=white] (-2.5,-9.5) rectangle ++(5,1);
        \node at (0,-8.75) {\textbf{preprocessed dataset}};
        \node at (0,-9.25) {$63 \times 134$};
    \end{tikzpicture}
    \caption{Preprocessing pipeline for the dataset.}
    \label{fig:preprocessing}
\end{figure}

The data preprocessing consists of several chronological steps prior to applying the ensemble feature selectors, see Fig. \ref{fig:preprocessing}.

\paragraph{Data cleaning} The first step in data preprocessing is to discard features known to be unimportant, such as features with only one unique value for all patients or duplicated features. Furthermore, we remove all data columns containing more than 25\% missing values across all patients. By this criterion, we remove 16 features from block (p), one feature from block (b), six features from block (h), 14 features from block (i), and eight features from block (t). 

Further, three patients are excluded from the experiments due to a high number of missing values in at least one block. All subsequent preprocessing steps are conducted on the remaining 63 patients and are applied by block to retain the homogeneous block structure.

\paragraph{Missing values}
Some values were missing because the clinicians did not fill out the case registration forms (CRF) properly or completely. Amongst other reasons, this may be because the information was missing in the patient journal, a blood sample was not done, a parameter was forgotten registered in the patient journal, or because the patients are referred from other hospitals. Such features, which are unavailable for a large percentage of patients, cannot be assessed properly in a data-driven manner and were therefore excluded --- an imputation of those features would be unreliable due to the small sample size and may introduce incorrect or misleading information into the model.

As a second step, we impute the features with less than 25\% missing values via an adaptation of the $k$-nearest neighbors ($k$NN) imputation algorithm~\cite{kuhn2013applied}. The number of features and the number of patients that have at least one missing value for each block are: (p) (7:25), (b) (5:16), (h) (7:6), (i) (2:2), and (t) (3:3) where the first number represents the number of features and the second number represents the number of patients.

In particular, we restrict the feature space to non-missing columns and compute a matrix of pair-wise distances between all patients. We denote the set comprising the $k$-nearest neighbors of patient $i$ by $N_k(i)\subseteq \{1,\dots,m\}$. Assuming that feature $j$ is missing for patient $i$, we impute $x_{i,j}$ by $x_{i,j}^{\text{imp}}$, representing the median (instead of the mean, as suggested by \cite{kuhn2013applied}) of feature $j$ across the patient's $k$ nearest neighbors where the feature value is known, i.e.
\begin{equation}
    x_{i,j}^{\text{imp}} \leftarrow \text{median}\left\{x_{l,j}: l\in N_k(i)\right\}.
\end{equation}
Ordered categorical features are transformed to an integer scale before interpolation. The usage of an odd value of $k$ (by default, we use $k=5$) guarantees that the median returns an integer, which is a clear benefit over the mean when using the technique for ordered features.

\paragraph{Categorical feature encoding}
Categorical features require encoding in order to be processed alongside numeric variables in predictive models.  In particular, we distinguish between ordinal and nominal categorical variables: Nominal variables (i.e., variables without an internal order of the feature levels), such as \textit{clinical institution}, are one-hot encoded~\cite{zheng2018feature}. Given a feature $j$ with $c_j$ feature levels, the one-hot encoding produces a set of $c_j-1$ binary features $\{\bm{e}_2,\dots,\bm{e}_{c_j}\}$, given as follows:
\begin{equation}
    \left(e_l\right)_i = \left\{\begin{array}{ll} 1 & \text{if}~x_{i,j} = l, \\ 0 & \text{otherwise,} \end{array}\right.
\end{equation} 
for $l\in \{2,\dots,c_j\}$ indicating the feature level.
The number of one-hot/ordinal categorical features is:  21/5 for block (p), 0/3 for block (b), 10/2 for block (h), 1/0 for block (i), and 5/0 for block (t).
To avoid linear dependencies between features, the first feature level is not represented by a binary vector in the encoded space, but rather contributes to the model intercept, see Tab.~\ref{tab:encodings}.

Features with an internal order among their levels (ordinal variables), such as the \textit{WHO performance status} with levels \textit{0, 1, 2, 3,} and \textit{4}, require an ordinal encoding to retain the relevant information about the order. Under the assumption that the influence of a feature increases from lower to higher levels (i.e., higher levels comprise the lower levels and an additive effect), the following encoding is used:
\begin{equation}
    \left(e_l\right)_i = \left\{\begin{array}{ll} 1 & \text{if}~x_{i,j} \leq l, \\ 0 & \text{otherwise.} \end{array}\right.    
\end{equation}
for feature level $l\in \{2,\dots,c_j\}$.
Again, the first feature level, which would be assigned a value of $1$ across all samples in the encoded space, is not assigned a binary vector in the encoded space. A comparison between one-hot and ordinal encoding is provided in Tab. \ref{tab:encodings}. In contrast to transforming to an integer scale, this binary ordinal encoding preserves the order among the categories but does not pretend equal distances between the categories on a numerical scale.

\begin{table}
    \centering
    \caption{One-hot versus ordinal encoding of a 4-level variable (levels A, B, C, D). Ordinal encoding assumes an order of the levels (here: A$<$B$<$C$<$D).}
    \label{tab:encodings}
    \begin{tabular}{c|c|c}
        \toprule
        level & one-hot encoding & ordinal encoding \\
        \midrule
        A & (0,0,0) & (0,0,0) \\
        B & (0,0,1) & (0,0,1) \\
        C & (0,1,0) & (0,1,1) \\
        D & (1,0,0) & (1,1,1) \\
        \bottomrule
    \end{tabular}
\end{table}

\paragraph{Feature transformation and normalization}
During our experiments, we split the dataset into train and test sets. To normalize the distribution of each numeric feature, we use the Yeo-Johnson power transformation along with standardization~\cite{yeo2000transform}. The Yeo-Johnson power transformation is an extension of the well-established Box-Cox transformation with the benefit that it enables the transformation of negative and zero values. The intention is to bring the data closer to a normal distribution by simultaneously stabilizing data variance. For a given feature $j$, Yeo-Johnson's power transform is defined as 

\begin{equation}
    x_{i,j}^{\text{YJ}} \leftarrow
    \begin{dcases}
    \frac{((x_{i,j} + 1)^{\lambda_j} -1)}{\lambda_j} & \text{if } \lambda_j \neq 0, x_{i,j}\geq 0 \\
    \log(x_{i,j}+1) & \text{if } \lambda_j = 0, x_{i,j}\geq 0 \\
    -\frac{((-x_{i,j}+1)^{2-\lambda_j}-1)}{2-\lambda_j} & \text{if } \lambda_j \neq 2, x_{i,j}<0 \\
    -\log(-x_{i,j}+1) & \text{if } \lambda_j = 2, x_{i,j}<0. \\
    \end{dcases}
\end{equation}

Commonly, the transformation parameter $\lambda_j$ is estimated from the data using a maximum likelihood approach. After the Yeo-Johnson transformation, we scale the data to zero mean and variance of 1. To prevent biased train and test data, the transformation parameter $\lambda_j$ and the mean and variance for the standardization are estimated on the training data in each split separately.

\paragraph{Encoding of the target variable}
Even though machine learning models for censored data are evolving, most present predictive models cannot handle censored data \cite{survival_ch1_1}. To avoid the problem presented by censored data, we encode the OS in months into an integer value (1-6). Using 60 months median follow-up time as a reference, there are no censored patients with OS below 60 months. Considering survival on a yearly basis we use the representation of the target variable in our experiments as in Tab. \ref{tab:encoding}. Since each level in the encoded space equals one year, predictive errors used in the remainder of this paper refer to a yearly scale.

\begin{table}
    \centering
    \caption{Encoding of the target variable "overall survival" (OS) [months].}
    \label{tab:encoding}
    \begin{tabular}{l|c}
        \toprule
         level & encoding\\
         \midrule
         OS $\leq 12$ & 1 \\
         $12<$ OS $\leq 24$ & 2 \\
         $24<$ OS $\leq 36$ & 3 \\
         $36<$ OS $\leq 48$ & 4 \\
         $48<$ OS $\leq 60$ & 5 \\
         $60<$ OS & 6 \\
         \bottomrule
    \end{tabular}
\end{table}

\subsection{Feature Selection Methods}
\begin{figure*}[h]
    \centering
    \begin{tikzpicture}[scale=0.8]
            \draw (0,0) rectangle ++(1,1);
            \node at (0.5,0.5) {$\bm{X}$};
            \node[align=center] at (-1.25,0.5) {input \\ dataset};
            
            \draw[->] (0.5,0) -- (-1,-1.5);
            \draw[->] (0.5,0) -- (0.5,-1.5);
            \draw[->] (0.5,0) -- (2.5,-1.5);
            \node[align = center] at (2.75,-0.5) {sub- \\ sampling};
            
            \draw (-1.5,-2.5) rectangle ++(1,1);
            \draw (0,-2.5) rectangle ++(1,1);
            \draw (2,-2.5) rectangle ++(1,1);
            \node at (-1,-2) {$\bm{X}_1$};
            \node at (0.5,-2) {$\bm{X}_2$};
            \node at (1.5,-2) {$\dots$};
            \node at (2.5,-2) {$\bm{X}_m$};
            \node[align=center] at (-2.75,-2) {data \\ subsets};
            
            \draw[->] (-1,-2.5) -- (-1,-4);
            \draw[->] (0.5,-2.5) -- (0.5,-4);
            \draw[->] (2.5,-2.5) -- (2.5,-4);
            \node[align=center] at (3.75,-3.25) {elementary \\ feature \\ selectors};
    
            \draw[fill=yellow!60] (-1.5,-5) rectangle ++(1,1);
            \draw[fill=yellow!60] (0,-5) rectangle ++(1,1);
            \draw[fill=yellow!60] (2,-5) rectangle ++(1,1);
            \node at (-1,-4.5) {$\bm{\delta}_1$};
            \node at (0.5,-4.5) {$\bm{\delta}_2$};
            \node at (1.5,-4.5) {$\dots$};
            \node at (2.5,-4.5) {$\bm{\delta}_m$};
            \node[align=center] at (-2.75,-4.5) {elementary \\ feature sets};
    
            \draw[->] (-1,-5) -- (0.5,-7.5);
            \draw[->] (0.5,-5) -- (0.5,-7.5);
            \draw[->] (2,-5) -- (0.5,-7.5);
            \draw[->] (0.5,-7.5) -- (0.5,-9.5);
            \node[align=center,fill=yellow!60] at (0.5,-5.75) {\textit{information from data}};

            \draw[dashed] (0.5,-7.5) -- (6,-7.5);
            \draw[dashed] (0.5,-7.5) -- (-6,-7.5);
            \draw[fill=blue!20,dashed] (0.5,-7.5) ellipse (2 and 1);
            \node[align = center] at (0.5,-7.5) {meta-model};

            \draw[fill=blue!20] (-5, -8) ellipse (2 and 1);
            \node at (-3, -7) {\textbf{RENT}};
            \node at (-5,-7.5) {selection criteria};
            \node at (-6.25, -8.5) {$\tau_1$};
            \node at (-5, -8.5) {$\tau_2$};
            \node at (-3.75, -8.5) {$\tau_3$};

            \draw[->] (6,-4.5) -- (6,-8);
            \draw[->] (8.25,-4.5) -- (6,-8);
            \node at (4, -7) {\textbf{UBayFS}};
            \draw[fill=blue!20] (6, -8) ellipse (2 and 1);
            \node[align=center] at (6,-8) {posterior distribution \\ over $\bm{\delta}$};
            \draw[fill=green!30] (5,-5) rectangle ++(2,1);
            \draw[fill=green!30] (7.25,-5) rectangle ++(2,1);
            \node[align=center] at (6,-4.5) {prior \\ knowledge};
            \node[align=center] at (8.25,-4.5) {side \\ constraints};
            \node[align=center,fill=green!30] at (7.25,-5.75) {\textit{information from expert}};

            \draw (0,-10.5) rectangle ++(1,1);
            \node at (0.5,-10) {$\bm{\delta}^{\star}$};
            \node[align=center] at (-1.25,-10) {final \\ feature set};
    \end{tikzpicture}
    \caption{Overall structure of both ensemble feature selection methods, RENT and UBayFS. After training elementary feature selectors, information is combined in a meta-model. While RENT uses information from data only, UBayFS additionally includes expert information.}
    \label{fig:meta_models}
\end{figure*}
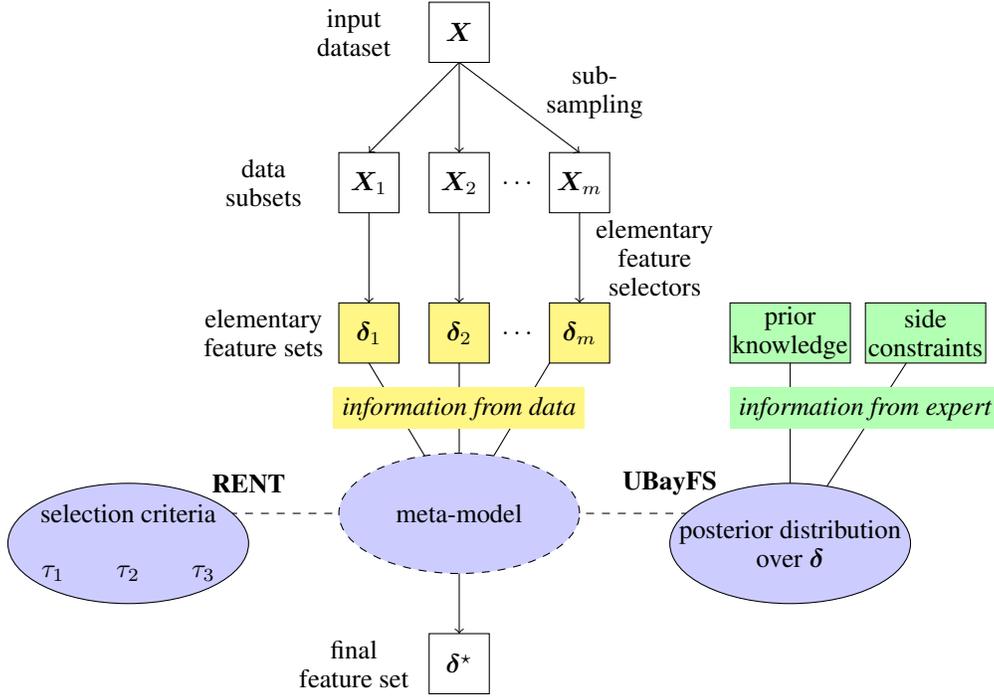

In this work, we investigate two ensemble feature selection methods, which have been tailored to fit the requirements of datasets in the life science domain: the Repeated Elastic Net Technique for Feature Selection (RENT)~\cite{jenul2021rent} and the User-Guided Bayesian Framework for Feature Selection (UBayFS)~\cite{jenul2022ubayfs}. Both methods build on the principle of (a) randomly subsampling the input dataset and (b) training an elementary feature selection model on each sample. The final feature set is determined by applying a meta-model on the feature sets selected by the elementary models, see Fig. \ref{fig:meta_models}. In the case of RENT, the elementary feature selector type is restricted to elastic net regularization~\cite{zou2005elasticnet} using logistic regression models for binary classification problems or ordinary least squares linear regression models for regression problems, while UBayFS operates on an arbitrary elementary model type.

\paragraph{RENT} The rules to obtain a final feature set further demonstrate the distinct scopes of the methods: RENT defines three criteria $\tau_1$, $\tau_2$ and $\tau_3$ for the selection of features based on the distribution of their weights across the elementary models; (I) the number of times the feature weights are non-zero ($\tau_1$) is above a level specified by the user; (II) the alternation of the sign of the feature weights does not surpass a user-defined level ($\tau_2$); (III) the size of the feature weights deviate significantly from 0 ($\tau_3$). The hyperparameters for RENT comprise of a number $M$ of elementary models, an internal data split ratio, two parameters associated with the elastic net regularization in the elementary models ($C$ and $\ell_1$), as well as one cut-off parameter for each of the three criteria $\tau_1,\tau_2,\tau_3$.

\paragraph{UBayFS} In contrast, UBayFS combines the selection frequency of each feature across the elementary models with prior information from domain experts, along with side constraints. In particular, the prior weighting of features is possible, along with the definition of linear side constraints between features (and feature blocks). In practice, weights can represent knowledge about the importance of features, which is verified from previous publications. Side constraints enable the user to restrict the feature set's maximum size $\max_s$ and account for the intrinsic block structure during feature selection (e.g., in multi-source datasets). Hence, RENT implements a purely data-driven approach based on Elastic Net, while UBayFS is a general meta-model with capabilities to integrate contextual information about the data generation process. In its most basic setup, UBayFS requires as hyperparameters a number of elementary models $M$ and an internal data split ratio, a maximum number of features $\max_s$, and a model type to use as the elementary feature selector.

\subsection{Outcome prediction}

\paragraph{Linear regression} Given a set of selected features $S$, we make use of linear regression models~\cite{hastie2009statisticallearning} to model the target variable $\bm{y}$. In its simplest form, the linear regression model (with intercept) is given as 
\begin{equation}
    \bm{y} = \tilde{\bm{X}}\bm{\beta} + \bm{\varepsilon},
\end{equation}
where $\bm{\beta}\in \mathbb{R}^{n+1}$ is the model parameter vector, $\tilde{\bm{X}}$ denotes the matrix containing one column of ones, followed by the sub-matrix of $\bm{X}$ restricted to the columns contained in $S$. Further, $\bm{\varepsilon}\underset{\text{iid}}{\sim} N(0,\sigma^2)$ denotes the model error with constant error variance $\sigma>0$. By default, parameters of linear regression models are obtained via ordinary least squares (OLS), i.e. by minimizing the least squares error 
\begin{equation}
    \underset{\bm{\beta}}{\min} ~ \Vert \bm{y} - \tilde{\bm{X}}\bm{\beta}\Vert_2^2.
    \label{eq:ols_reg}
\end{equation}

Once the parameter vector $\bm{\beta}$ is estimated by optimizing Eq. \ref{eq:ols_reg} analytically, predictions are obtained by evaluating $\hat{\bm{y}} = \tilde{\bm{X}}\bm{\beta}$.

\paragraph{$k$-nearest neighbor ($k$NN) regression} As an alternative to the linear regression model, a $k$-nearest neighbor ($k$NN) regression model~\cite{hastie2009statisticallearning} is used to compute predictive results. In contrast to the linear regression model, the $k$NN model does not assume a linear relationship between the predictors and the target variable. Similar to the $k$NN method used for missing value imputation in Section \ref{sec:preprocessing}, a neighborhood $N_k(i)$ of sample $i$ containing the $k$ nearest training data points with respect to a Euclidean metric on the feature space is computed for any data point $\bm{x}_i$. The prediction for the target value $y_i$ corresponding to sample $i$ is given by the mean of the neighbor's target values
\begin{equation}
    \hat{y}_i = \frac{1}{k}\sum\limits_{l\in N_k(i)} y_l.
\end{equation} 
Note that, the neighborhood $N_k(i)$ is a subset of the training samples only, while $\hat{y}_i$ may represent both, training or test samples.

Both predictive models, linear regression as well as the $k$NN regression model, are known to suffer from the curse of dimensionality --- hence, we can assume that selecting a high number of features deteriorates each model. The opposite extreme for both methods, i.e., selecting no features at all, leads to predicting the output with the mean over the training data regardless of the input. Thus, we expect a well-performing feature selector to deliver a proper subset $S$ of the feature set $\{1,\dots,n\}$, which allows both predictive models to perform better than the baselines given by (a) the overall mean of the target variable, and (b) a model including all features.

\subsection{Implementation}

Parts of our analyses are conducted in the programming languages \texttt{R}~\cite{r2022}; other parts are conducted in Python~\cite{python2009}. We use the open-source implementations for RENT~\cite{jenul2021rent_software} and UBayFS~\cite{jenul2023ubay_software}. For data preparation and preprocessing, we deploy the R package \textit{caret}~\cite{caret2022}, and the Python package \textit{scikit-learn}~\cite{scikit-learn2011}. Fold indices are shared between \texttt{R} and Python. Predictive models are trained and evaluated in \texttt{R} using the caret package for all model setups. All plots are created using package \textit{ggplot2}~\cite{ggplot2016}.

All results are produced on an Intel Core i7 CPU @1.8 GHz, 32GB RAM under a Windows 11 Pro operating system.

\section{Experiments}
\label{sec:experiments}
Our experimental results are structured into a pre-study, where we determine optimal hyperparameters for the feature selection algorithms, followed by two main experiments. Experiment 1 focuses on the comparison of the two models, RENT and UBayFS, on the dataset without accounting for additional expert knowledge. Experiment 2 is operated on UBayFS only, as prior information and additional side constraints are included in the feature selection.

Our main focus in the experiments lies on the selected feature sets, along with the impact of the feature selection on predictive performance. We provide feature counts from both of the investigated feature selectors, RENT and UBayFS, across five different train-test splits of the dataset. Unless specified otherwise, all experiments are conducted using the hyperparameters determined during the pre-study.

\subsection{Experimental setup}

\paragraph{Model parameters}
Both algorithms, RENT and UBayFS, are trained on $M=100$ ensemble models and internal 0.75/0.25-splits for sub-sampling the dataset. The underlying elementary feature selector for RENT is, by definition, an elastic net regularized linear regression model. Thus, RENT requires five hyperparameters to be determined during the pre-study (2 elastic net regularization parameters, $\ell_1$ and $C$, as well as three thresholds $\tau_1,\tau_2,$ and $\tau_3$ for the selection criteria). 
In order to make results comparable with UBayFS, we further deploy a side condition to restrict the search space to settings, which deliver a maximum number of features $\max_s$ during validation. Thus, the number of features selected by RENT is approximately equal to the pre-defined parameter $\max_s$.

UBayFS uses minimum redundancy max relevance (mRMR)~\cite{ding2005mrmr} as an elementary feature selector. The internal number of features in each elementary model is set to $\max_s$, i.e., each elementary model selects exactly $\max_s$ features. For the meta-model, the same parameter $\max_s$ is used to restrict the maximum number of selected features via a max-size side constraint (hard constraint) --- while different levels of $\text{max}_s$ are evaluated in experiment 1, the parameter is set to the default $\max_s = 20$ in experiment 2. Further, unless otherwise stated, prior feature weights in UBayFS are set uniformly to $0.1$ across all features, which results in a non-informative prior.

\paragraph{Train-test splits}
As the ratio between the number of patients and features is unbalanced, with 63 patients and 134 encoded features, the reliability of the feature ranking results must be validated to reduce the risk of spurious correlations and overfitting. Hence, we perform a 5-fold split of the dataset. For all possible permutations, we use four folds for training UBayFS or RENT, as well as the predictive models and the remaining fold for testing. Hyperparameters are determined on each split separately by internally subsetting the 4-fold training set (nested split). The 5-fold splits and hyperparameters determined in the pre-study remain the same across all experiments.

For each feature selection method, we provide the selection frequencies of each feature across the 5 folds, i.e., a feature obtains an importance score between 0 and 5 according to the number of folds it was selected for. For predictive performance scores, a linear regression model and a $k$NN regression model are trained on the same training folds, using the features from the preceding feature selection, and evaluate the prediction error on the test set (averaged across all folds).

\paragraph{Performance metrics}
To assess whether a feature set contains relevant information for training predictive models, we analyze the predictive performance in a regression setup following the feature selection step. The performance is quantified using the root mean squared error (RMSE), which has a lower bound of 0 and shall be minimized.

Using the stability criterion introduced by \cite{nogueira2018stability}, we further evaluate the feature selection stability across the five folds for RENT and UBayFS. The computed score is bounded in the interval $[0,1]$; a value of 1 indicates perfect stability, i.e., the same feature set is selected in each model, while 0 indicates that selected feature sets show no overlap. 

Furthermore, the redundancy rate (RED) returns an intrinsic feature set quality measure by computing the average absolute Pearson correlation among the selected features. Small correlations are desirable as highly correlated features represent redundant information. Equally to the absolute Pearson correlation coefficient, RED is bounded in $[0,1]$.

In experiment 2, we additionally assign prior weights to a subset of features --- therefore, we also evaluate the percentage of prior-elevated features (PERC) in the selected feature sets as well. If PERC is high, features extracted via data-driven feature selectors match the domain experts' knowledge. However, a low PERC does not necessarily contradict expert knowledge since the features may be highly correlated, and therefore, similar information may be encoded in multiple distinct sets of features.

\subsection{Pre-Study}
The pre-study aims to determine the optimal hyperparameters for RENT. Given a 0.75/0.25 outer train-test split as specified above, only train data are used for hyperparameter selection. For this purpose, 4-fold cross-validation is performed on each train dataset (using the same 4 folds as in the outer train-test split). Across the resulting 4 models, hyperparameters are selected by maximizing predictive performance in a grid search over the parameter space $C\in\{1,10,100,1000\}$, $\ell_1\in \{0,0.1,0.2,\dots,1\}$, $\tau_1,\tau_2\in \{0,0.05,0.1,\dots,1\}$, and $\tau_3=0.975$ (fixed).

The runtime for the full computation associated with the pre-study (parameter selection and final feature selection) for RENT comprised approx. 350 sec (16 cores, 24 threads in parallel). Since UBayFS does not require parameter selection, the runtime to evaluate the feature selection model for different levels of $\max_s$ (see Experiment 1) is shorter (approx. 65 sec without parallelization).

\begin{table}
\centering
\caption{Selected hyperparameters for each train/test split.}
\label{tab:pre-study}
\begin{tabular}{ll | lllll}
\toprule
& \multirow{2}{*}{parameter} & \multicolumn{5}{c}{fold} \\
&&  1 & 2 & 3 & 4 & 5 \\  \midrule 
\multirow{5}{*}{RENT} & $\ell_1$ & 0.3 & 0 & 0.3 & 0.3 & 0.3 \\
&$C$ & 1 & 1 & 1 & 1 & 1 \\
&$\tau_1$ & 0.3 & 0.5 & 0.3 & 0.35 & 0.35 \\
&$\tau_2$ & 0.3 & 0.5 & 0.4 & 0.35 & 0.35 \\ 
&$\tau_3$ & 0.975 & 0.975 & 0.975 & 0.975 &  0.975 \\ \midrule
UBayFS & $\max_s$ & 20 & 20 & 20 & 20 & 20 \\ \bottomrule
\end{tabular}
\end{table}

Tab. \ref{tab:pre-study} shows the hyperparameters identified for RENT and UBayFS in each train-test split (given by the numbers of the test folds 1-5). Due to the restriction of the maximum number of features, the stated parameters may not represent global maxima for the performance of RENT; however, comparability between the methods is preserved. Furthermore, since the number of features is restricted, the selected hyperparameters are in a similar range between the folds.

\subsection{Experiment 1: feature selection without prior knowledge}

Having determined hyperparameters for each fold in the pre-study, RENT, and UBayFS are applied in each data split to the training dataset to select an optimized feature set for a given $\max_s$ on a purely data-driven basis.

\paragraph{Selected features} For each feature, selection frequencies across the five test folds are further provided in Tab. \ref{tab:feature_ranking_exp1} (columns \textit{RENT} and \text{UBayFS, $w=0.1$}). In addition to the selection frequency, the table indicates whether a feature shows a positive or negative impact on the target variable according to the coefficients in the linear model, if selected. Thereby, $++$ and $-{}-$ indicate that a feature always shows the same sign across all predictive models. In contrast, $+$ and $-$ indicate a majority of positive or negative coefficients across the predictive models, respectively.

\begin{landscape}
\begin{table}[t]
\centering
\caption{Feature selection frequencies across five folds by RENT and UBayFS (with different prior weight levels $w$ for selected features). Features with increased prior weights in the UBayFS setup reported in the last column are highlighted with asterisks. In parentheses, $+$ and $-$ indicate that the majority of linear regression models containing the feature assigned it a positive or negative coefficient, respectively; $++$ or $-{}-$ indicate that all models containing the feature had equal signs of their coefficients, while no parentheses indicate that the distribution of signs was even.}
\label{tab:feature_ranking_exp1}
\begin{minipage}{0.48\textwidth}
\centering
\resizebox{\textwidth}{!}{
\begin{tabular}{lllrrrr}
  \toprule
\multirow{2}{*}{block} & & \multirow{2}{*}{feature} & \multirow{2}{*}{RENT} & \multicolumn{3}{c}{UBayFS} \\ 
& & & & $w=0.1$ & $w=50$ & $w=110$ \\
  \midrule
  \multirow{39}{*}{(p)}
   & * & Age at Diagnosis & 2 & 1(-{}-) & 5(-) & 5(-{}-) \\ 
   &  & Time from PET to Metastasis (days)  & 0 & 1(++) & 1(++) & 0 \\ 
   &  & Time from PET to Diagnosis  (days)  & 0 & 0 & 0 & 0 \\ 
   &  & Time from diag to mets  (months)  & 0 & 0 & 0 & 0 \\ 
   &  & Sex & 0 & 0 & 0 & 0 \\ 
   &  & Loc. Adv. Resectable Disease & 0 & 0 & 0 & 0 \\ 
   &  & Loc. Reccurence & 0 & 0 & 0 & 0 \\ 
   &  & Metastatic Disease at Time of Diagnosis & 3(+) & 1(++) & 0 & 0 \\ 
   &  & Treatment Intention Palliative & 4(-) & 5(-{}-) & 3(-{}-) & 0 \\ 
   &  & Prior Other Cancer & 2(++) & 1(-{}-) & 0 & 0 \\ 
   &  & Living Alone & 0 & 0 & 0 & 0 \\ 
   & * & TNM staging Pathological & 0 & 0 & 0 & 5(-) \\ 
   &  & Stage grouped Stage IV & 0 & 0 & 0 & 0 \\ 
   &  & Mets Bone & 5(-{}-) & 5(-{}-) & 5(-{}-) & 0 \\ 
   &  & Mets LN Distant & 0 & 0 & 0 & 0 \\ 
   &  & Mets LN Regional & 0 & 0 & 0 & 0 \\ 
   &  & Mets LN Retro & 0 & 0 & 0 & 0 \\ 
   &  & Mets LN & 0 & 0 & 0 & 0 \\ 
   &  & Mets Liver & 0 & 0 & 0 & 0 \\ 
   &  & Mets Lung & 0 & 0 & 0 & 0 \\ 
   &  & Mets Other & 0 & 0 & 0 & 0 \\ 
   &  & Mets Skin & 0 & 0 & 0 & 0 \\ 
   &  & Primary Tumour Resected & 0 & 0 & 0 & 0 \\ 
   &  & M-stage M1 & 0 & 0 & 0 & 0 \\ 
   &  & BMI & 1(-{}-) & 0 & 0 & 0 \\ 
   &  & Non Smoker & 0 & 0 & 0 & 0 \\ 
   &  & Smoker & 0 & 1(++) & 0 & 0 \\ 
   &  & Radical Surgery & 3(++) & 4(+) & 4(+) & 0 \\ 
   &  & Co-morbidity Severity 1 & 0 & 0 & 0 & 0 \\ 
   &  & Co-morbidity Severity  $>$ 1 & 0 & 0 & 0 & 0 \\ 
   &  & T-stage T2 & 0 & 0 & 0 & 0 \\ 
   &  & T-stage T3 & 0 & 0 & 0 & 0 \\ 
   &  & T-stage T4 & 2(-{}-) & 2(-{}-) & 1(-{}-) & 0 \\ 
   &  & N-stage N1 & 0 & 0 & 0 & 0 \\ 
   &  & N-stage $>$  N1 & 0 & 0 & 0 & 0 \\ 
   & * & WHO Perf Stat 1 & 0 & 0 & 3(-{}-) & 5(+) \\ 
   & * & WHO Perf Stat 2 & 4(-{}-) & 5(-{}-) & 5(-{}-) & 5(-{}-) \\ 
   & * & WHO Perf Stat 3 & 0 & 0 & 0 & 3(-{}-) \\ 
   & * & WHO Perf Stat 4 & 0 & 0 & 0 & 3(-) \\
   \midrule
   \multirow{15}{*}{(b)}
   &  & Abs.  Neutrophil Count & 0 & 0 & 0 & 0 \\ 
   & * & Albumin & 2 & 4(-{}-) & 5(-) & 5(+) \\ 
   &  & CRP & 5(-) & 5(-{}-) & 5(-) & 0 \\ 
   &  & Creatinine & 0 & 0 & 0 & 0 \\ 
   &  & Haemoglobin & 0 & 0 & 0 & 0 \\ 
   &  & WBC & 1(-{}-) & 1(-{}-) & 1(-{}-) & 0 \\ 
   &  & ALP $>$ Normal $<$= 3UNL & 4(-{}-) & 5(-{}-) & 4(-{}-) & 0 \\ 
   &  & ALP $>$ 3UNL & 1(++) & 0 & 0 & 0 \\ 
   &  & Chromogranin\_A $>$ Normal $<$= 2UNL & 0 & 0 & 0 & 0 \\ 
   &  & Chromogranin\_A $>$ 2UNL & 0 & 0 & 0 & 0 \\ 
   & * & LDH $>$ Normal  $<$= 2UNL & 0 & 0 & 1(++) & 4(++) \\ 
   & * & LDH $>$ 2UNL & 0 & 0 & 1(++) & 5(+) \\ 
   &  & NSE $>$ Normal $<$= 2UNL & 0 & 0 & 0 & 0 \\ 
   &  & NSE $>$ 2UNL & 0 & 0 & 0 & 0 \\ 
   & * & Platelets & 2(-{}-) & 4(-{}-) & 5(-{}-) & 5(-{}-) \\ 
   \midrule
      \end{tabular}}
    \end{minipage}
    \hskip15pt
\begin{minipage}{0.48\textwidth}
\centering
\resizebox{\textwidth}{!}{
\begin{tabular}{lllrrrr}
    \toprule
    \multirow{2}{*}{block} & & \multirow{2}{*}{feature} & \multirow{2}{*}{RENT} & \multicolumn{3}{c}{UBayFS} \\ 
& & & & $w=0.1$ & $w=50$ & $w=110$ \\
\midrule
   \multirow{30}{*}{(h)}
   & * & Ki-67 & 5(-{}-) & 5(-{}-) & 5(-{}-) & 5(-{}-) \\ 
   &  & Hist Exam Metastasis & 0 & 0 & 0 & 0 \\ 
   & * & Primary Tumour Esophagus & 0 & 0 & 0 & 3(-) \\ 
   & * & Primary Tumour Gallbladder/duct & 0 & 0 & 1(++) & 4(+) \\ 
   & * & Primary Tumour Gastric & 0 & 0 & 1(-{}-) & 5(-) \\ 
   & * & Primary Tumour Other abdominal & 0 & 0 & 0 & 5(-) \\ 
   & * & Primary Tumour Pancreas & 1(++) & 0 & 1(++) & 5(++) \\ 
   & * & Primary Tumour Rectum & 0 & 0 & 0 & 5(++) \\ 
   & * & Unknown Pr. With Dominance of GI met. & 0 & 0 & 0 & 3(+) \\ 
   &  & Co-existing Neoplasm Adenoma & 0 & 0 & 0 & 0 \\ 
   &  & Co-existing Neoplasm Dysplasia & 0 & 0 & 0 & 0 \\ 
   &  & No Co-existing Neoplasm & 0 & 0 & 0 & 0 \\ 
   & * & Tumour Morphology WD & 4(+) & 3(-{}-) & 4 & 5(-{}-) \\ 
   &  & Chromogranin A Staining & 0 & 0 & 0 & 0 \\ 
   &  & Architecture Infiltrative & 1(++) & 0 & 0 & 0 \\ 
   &  & Architecture Organoid & 1(++) & 1(++) & 0 & 0 \\ 
   &  & Architecture Solid & 0 & 0 & 0 & 0 \\ 
   &  & Architecture Trabecular & 1(-{}-) & 1(++) & 0 & 0 \\ 
   &  & Vessel Pattern Distant & 1(++) & 2(-{}-) & 1(-{}-) & 0 \\ 
   &  & Biopsy Location Gastric & 0 & 0 & 0 & 0 \\ 
   &  & Biopsy Location Liver Metastasis & 0 & 0 & 0 & 0 \\ 
   &  & Biopsy Location Lymph Node & 0 & 0 & 0 & 0 \\ 
   &  & Biopsy Location  Oesophagus & 0 & 0 & 0 & 0 \\ 
   &  & Biopsy Location Pancreas & 0 & 0 & 0 & 0 \\ 
   &  & Biopsy Location Peritoneum & 2 & 1(-{}-) & 0 & 0 \\ 
   &  & No Stroma & 4(++) & 1(++) & 1(++) & 0 \\ 
   &  & Stroma & 3(++) & 3(-) & 2(-{}-) & 0 \\ 
   &  & Geographic Necrosis & 0 & 1(-{}-) & 0 & 0 \\ 
   &  & Synaptophysin Staining 2+  & 0 & 0 & 0 & 0 \\ 
   &  & Synaptophysin Staining 3+ & 0 & 0 & 0 & 0 \\ 
   \midrule
   \multirow{12}{*}{(i)}
   &  & Injection to Scan [min] & 2 & 2(++) & 2(++) & 0 \\ 
   &  & Weight [kg] & 2(-{}-) & 0 & 0 & 0 \\ 
   & * & Total MTV [cmˆ3] & 3(+) & 1(-{}-) & 4(++) & 5(-{}-) \\ 
   &  & SUVmean & 0 & 0 & 0 & 0 \\ 
   & * & SUVmax & 2 & 4 & 5(++) & 5(-) \\ 
   &  & SUVmean (total) & 1(++) & 0 & 0 & 0 \\ 
   &  & SUVmax (total) & 5(-{}-) & 5(-{}-) & 5(-{}-) & 0 \\ 
   & * & Total TLG [g] & 4 & 1(++) & 3(+) & 5(++) \\ 
   &  & Institution Rikshospitalet & 4(++) & 4(++) & 2(++) & 0 \\ 
   &  & Institution Ullevaall & 0 & 0 & 0 & 0 \\ 
   &  & Height [cm] & 0 & 0 & 0 & 0 \\ 
   &  & Glucose [mmol/L] & 2(-{}-) & 0 & 0 & 0 \\ 
   \midrule
   \multirow{17}{*}{(t)}
   &  & Time from PET to first treatment  (days)  & 0 & 0 & 0 & 0 \\ 
   &  & Chemotherapy Type Cisplatin/Etoposide & 4(+) & 3(+) & 2(-{}-) & 0 \\ 
   &  & Chemotherapy Type Other & 0 & 0 & 0 & 0 \\ 
   &  & Chemotherapy Type Temozolomide/Capecitabine & 1(++) & 0 & 0 & 0 \\ 
   &  & Chemotherapy Type Temozolomide/Everolimus & 4(++) & 5(++) & 4(++) & 0 \\ 
   &  & Best Response  (RECIST) Not Assessed & 0 & 0 & 0 & 0 \\ 
   &  & Best Response  (RECIST) Only Clinical PD & 0 & 0 & 0 & 0 \\ 
   &  & Best Response  (RECIST) Partial Response & 2(-{}-) & 0 & 0 & 0 \\ 
   &  & Best Response  (RECIST) Progressive Disease  & 0 & 0 & 0 & 0 \\ 
   &  & Best Response  (RECIST) Stable Disease  & 0 & 0 & 0 & 0 \\ 
   &  & Reintroduction with Cisplatin Etoposide & 0 & 0 & 0 & 0 \\ 
   &  & Number of Courses & 4(++) & 4(++) & 3(++) & 0 \\ 
   &  & Treatment Stopped Other & 1(++) & 1(++) & 1(++) & 0 \\ 
   &  & Treatment Stopped Progression of Disease & 0 & 0 & 0 & 0 \\ 
   &  & Treatment Stopped Toxicity & 0 & 0 & 0 & 0 \\ 
   &  & No Progression & 5(++) & 4(++) & 3(++) & 0 \\ 
   &  & Progression & 3 & 3(+) & 1(-{}-) & 0 \\ 
   \bottomrule
\end{tabular}
}
\end{minipage}
\end{table}
\end{landscape}

\begin{figure}
     \centering
     \begin{subfigure}[b]{0.48\textwidth}
         \centering
         \includegraphics[width=\textwidth]{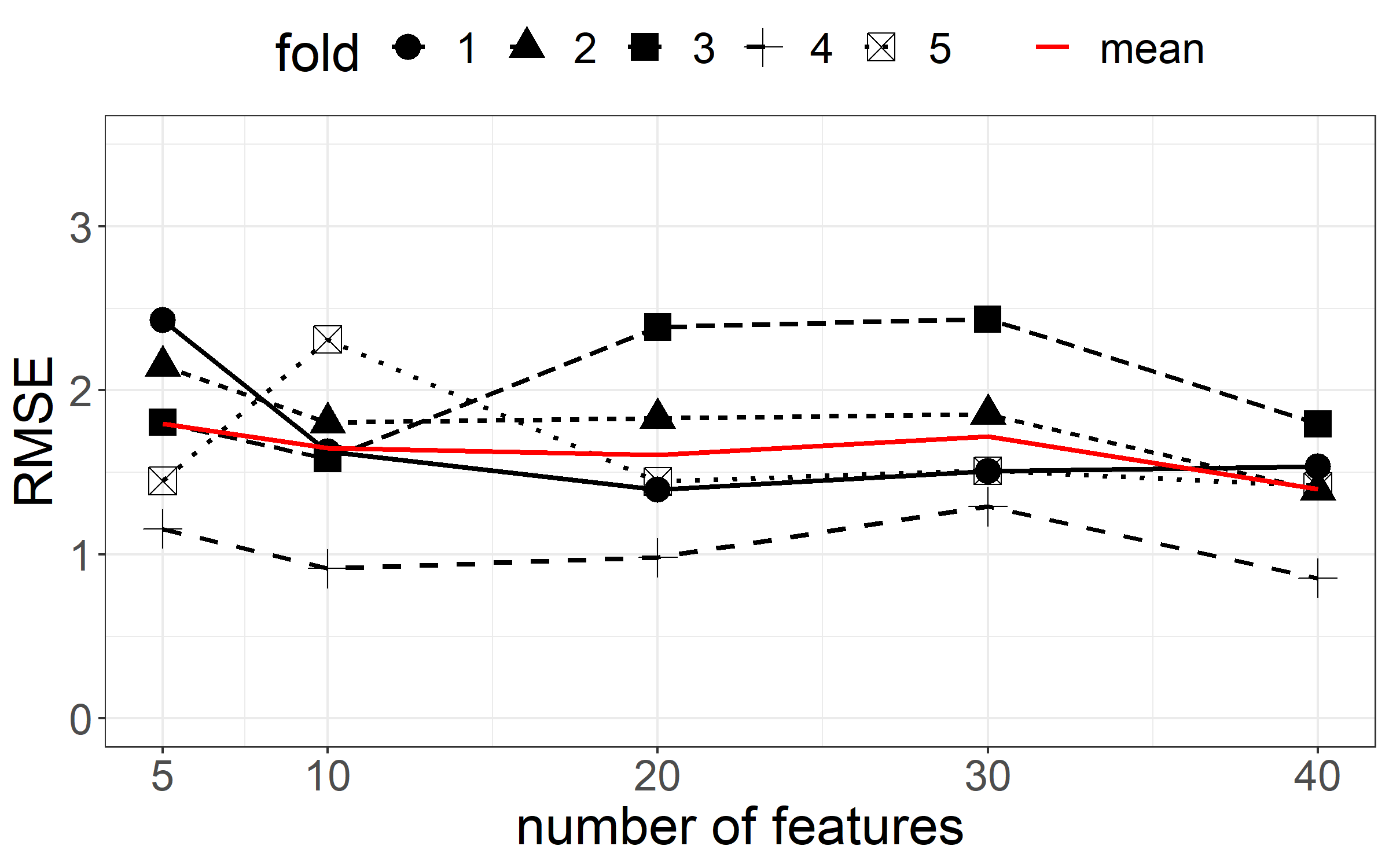}
         \caption{$k$NN regression on RENT features}
     \end{subfigure}
     \begin{subfigure}[b]{0.48\textwidth}
         \centering
         \includegraphics[width=\textwidth]{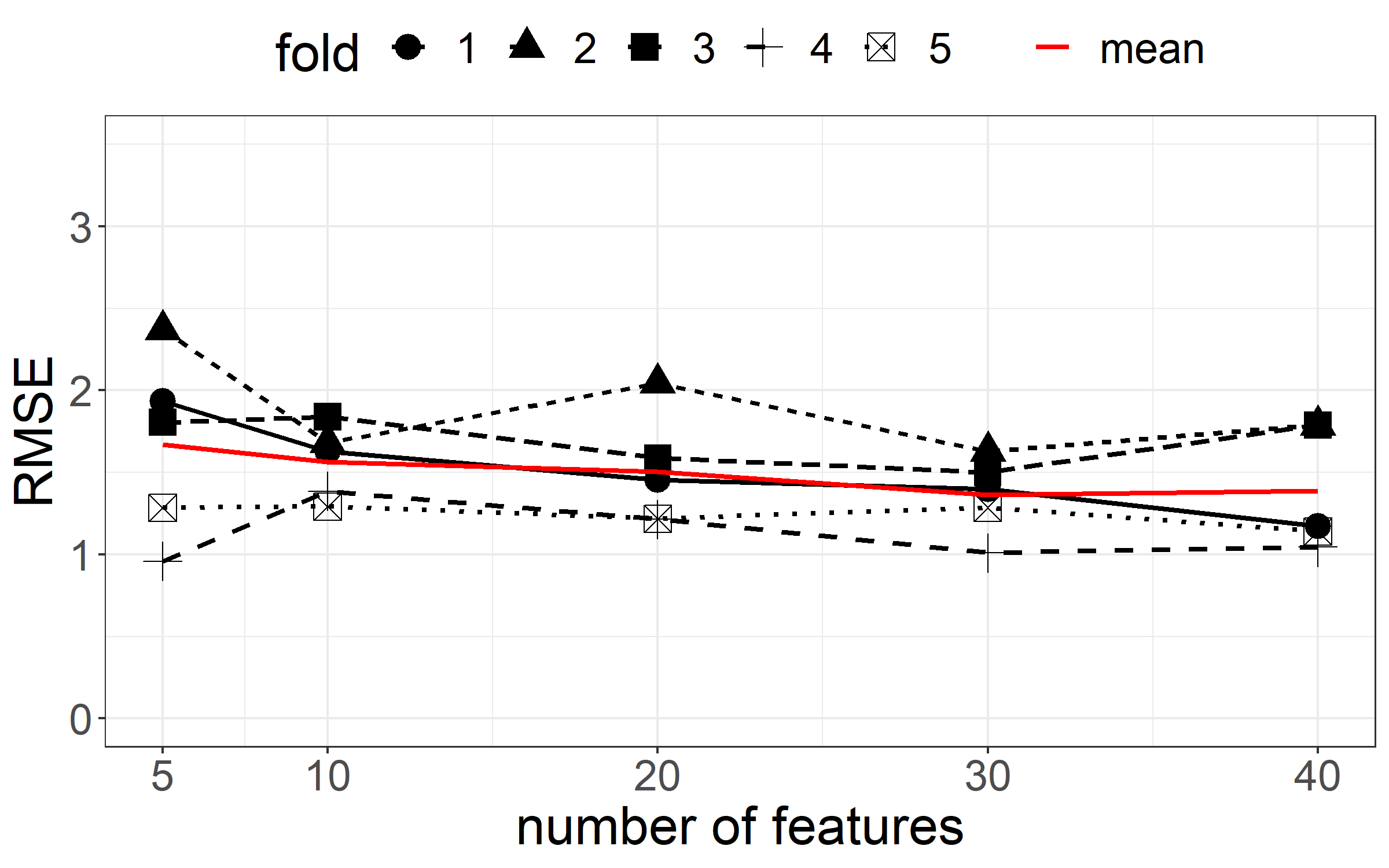}
         \caption{$k$NN regression on UBayFS features}
     \end{subfigure}
     \begin{subfigure}[b]{0.48\textwidth}
         \centering
         \includegraphics[width=\textwidth]{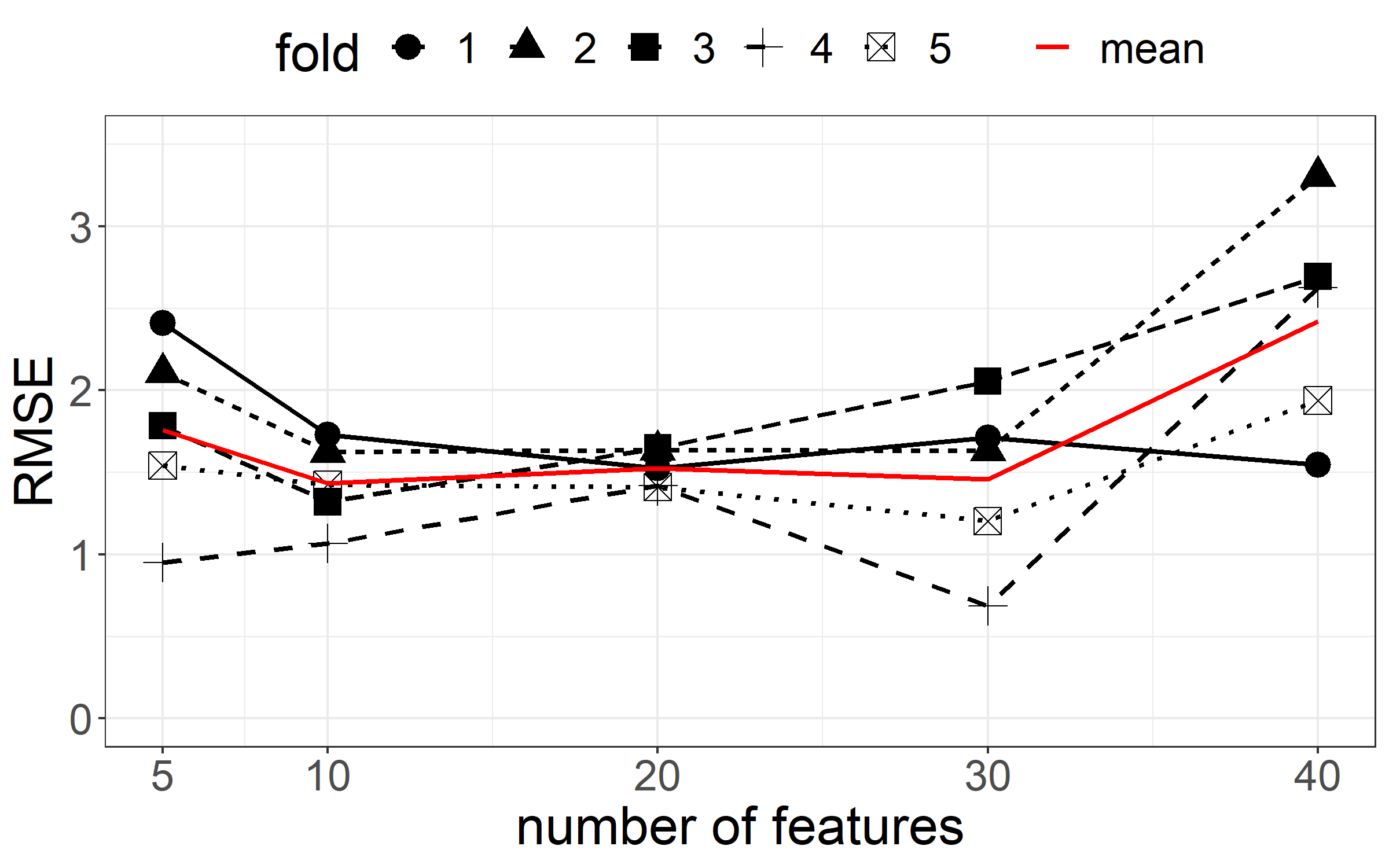}
         \caption{linear regression on RENT features}
     \end{subfigure}
     \begin{subfigure}[b]{0.48\textwidth}
         \centering
         \includegraphics[width=\textwidth]{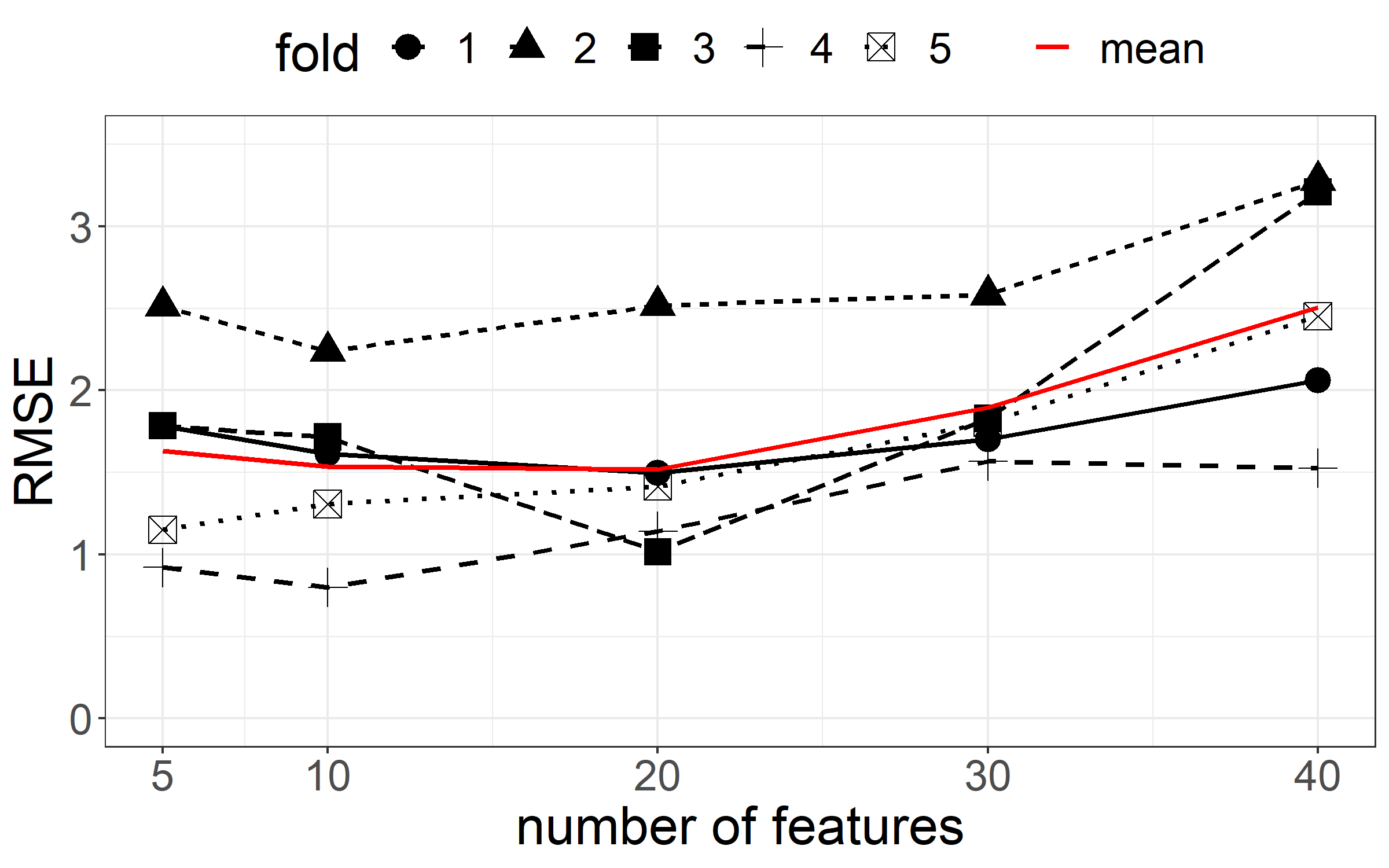}
         \caption{linear regression on UBayFS features}
     \end{subfigure}
    \caption{Predictive performances (on test set) of models trained after feature selection for different numbers of features.}
    \label{fig:performances_max_s_exp1}
\end{figure}

\paragraph{Predictive performance} Further, Fig. \ref{fig:performances_max_s_exp1} illustrates the predictive performances of $k$NN and linear regression models trained after UBayFS and RENT feature selection. The plot shows the RMSE for each fold given a predefined number of selected features $\max_s$.

Notably, RENT performs better using the linear regression model as the predictor, while UBayFS shows a better performance in combination with $k$NN. The stronger performance of RENT with linear regression may be a result of the fact that the underlying feature selection in RENT is based on a regularized linear regression model. UBayFS, however, is based on mRMR, which does not build upon a linear predictive model.

While linear regression results deteriorate at a higher number of features ($\max_s>30$), the $k$NN model retains a similar performance level, which suggests that the curse of dimensionality does not yet have a strong effect on the Euclidean distance for the given feature space dimensionalities. For the linear model, overfitting is triggered by a large ratio between the number of features and the number of patients. 

Among all compared methods, differences between the folds are obvious: for instance, fold 4 is predicted with the least RMSE across all combinations of feature selector, predictive model, and $\max_s$. On the other hand, fold 2 is associated with a large RMSE in the models based on UBayFS, while fold 3 shows similar behavior for the $k$NN model based on RENT features. Potentially, differences between folds may be caused by two factors (or combinations of both):
\begin{itemize}
    \item the cohort of patients in the \textit{training set} does not represent the global distribution of the data well --- e.g., the training data do not contain a sufficient number of samples with particularly high or low target values (bad prediction due to a bad model);
    \item the cohort of patients in the \textit{test set} is particularly hard to estimate, e.g., due to outliers (bad prediction in spite of an appropriate model);
\end{itemize}
Due to the low number of only 12-13 patients in each fold, even a low number of hard-to-predict outliers may deteriorate RMSE results significantly.

\begin{figure}
    \centering
    \includegraphics[width=0.6\textwidth]{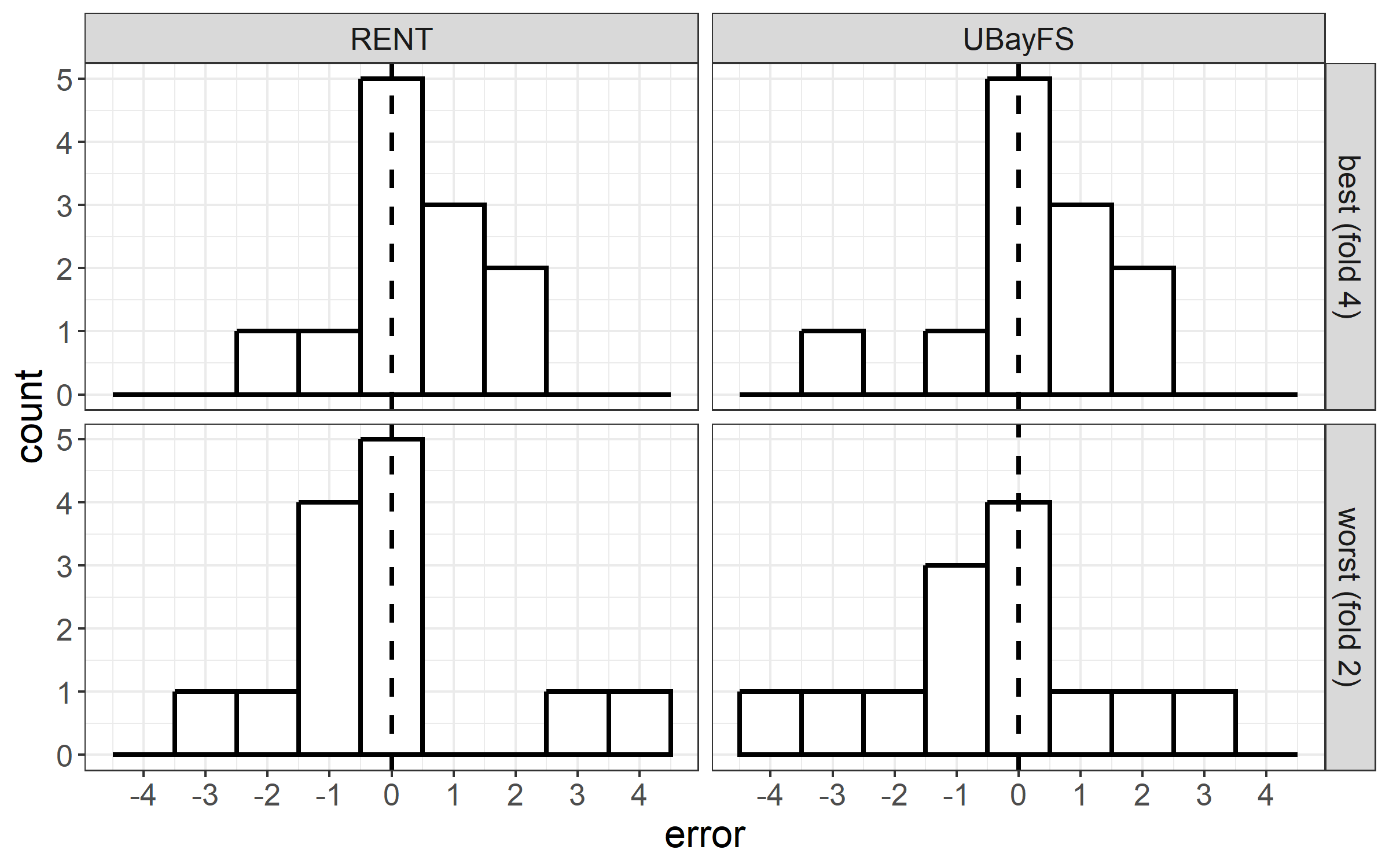}
    \caption{Histograms of errors on the test set (predicted value by $k$NN - ground truth) of the folds performing best (fold 4) and worst (fold 2) at $\max_s=20$ features.}
    \label{fig:histograms_exp1}
\end{figure}

\paragraph{Residuals} In order to shed light on the dynamics leading to the differences in performance between the data folds, histogram plots of the residuals for fold 2 (worst fold in UBayFS) and fold 4 (best fold across most setups) at $\max_s=20$ are provided in Fig. \ref{fig:histograms_exp1}. Residuals are defined as the difference between the true value and the prediction; thus a positive or negative residual value indicates an underestimation or overestimation of the lifetime, respectively.

In contrast to fold 4, the residuals from fold 2 are more dispersed. All histograms are symmetric and  centered around $0$, which indicates that all methods are able to estimate the intercept correctly. In both folds, the prediction model was able to predict the correct lifetime category for almost half of the patients in the test set. However, the histogram indicates that predictive models based on both feature selectors overestimate the lifetime in test fold 4 (positive errors), while lifetimes in test fold 2 are rather slightly overestimated (negative errors). The main difference in performance between fold 2 and fold 4 is driven by dispersion, i.e. by a minority of patients, which show a high error --- due to the small sample size, even a small number of such outliers can impact the total RMSE significantly. 

When considering patients with absolute residual values $>2.5$ as outliers, RENT, and UBayFS show 3 outliers in fold 2, each (RENT: 2 positive, 1 negative; UBayFS: 1 positive, 2 negative). Both methods commonly misclassify one patient with true target value 6 and predictions 2.6 (UBayFS) and 2 (RENT), which substantiates the highest positive outlier in both histograms. The remaining two outliers of each method refer to different patients.

\begin{figure}
     \centering
     \begin{subfigure}[b]{0.48\textwidth}
         \centering
         \includegraphics[width=\textwidth]{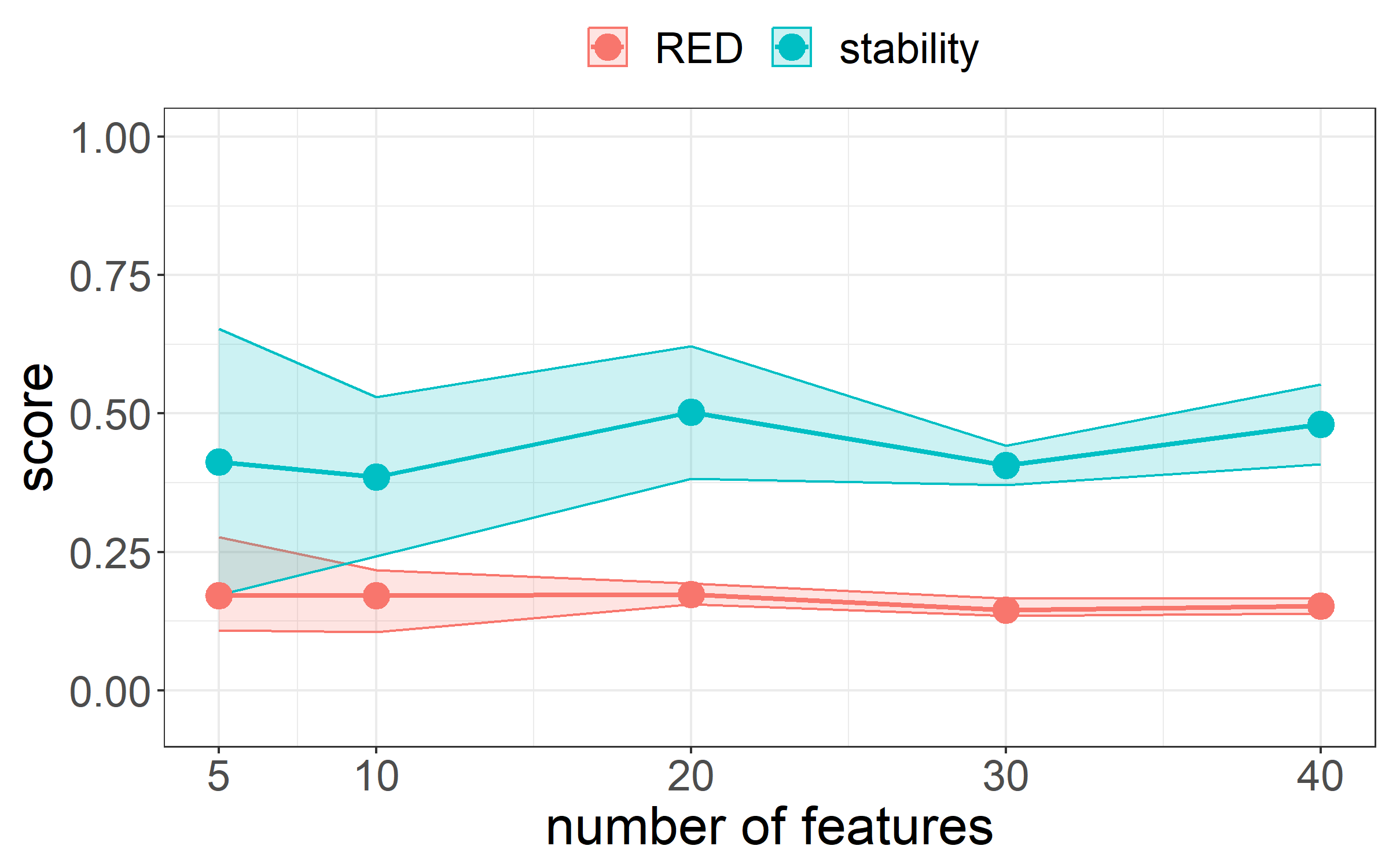}
         \caption{RENT features}
     \end{subfigure}
     \begin{subfigure}[b]{0.48\textwidth}
         \centering
         \includegraphics[width=\textwidth]{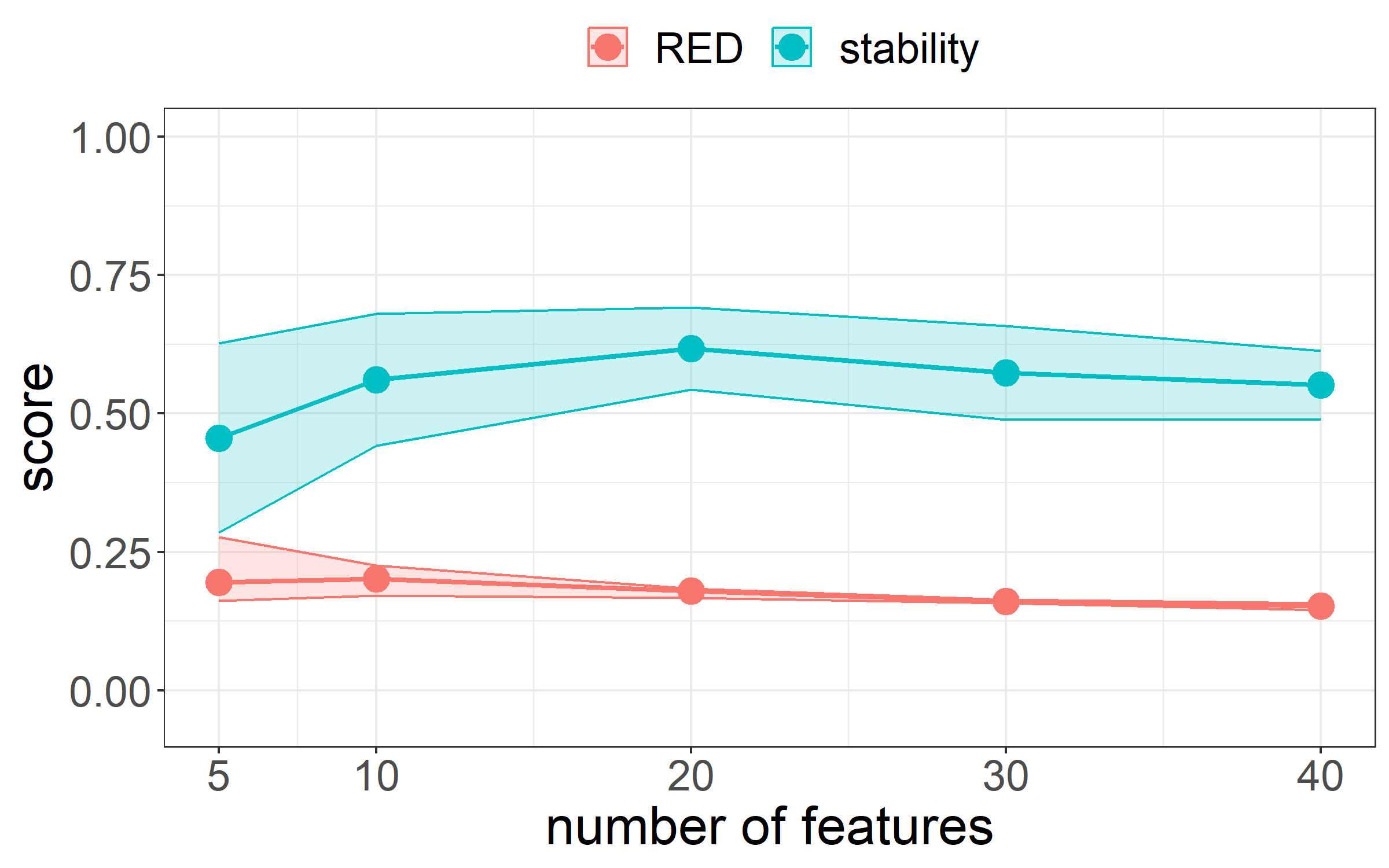}
         \caption{UBayFS features}
     \end{subfigure}
     \caption{Stabilities and redundancy rates (RED) of feature sets selected by RENT and UBayFS ($\max_s=20$ features, each).}
    \label{fig:stabilities}
\end{figure}

\paragraph{Stability} In addition to the performance evaluation, we further investigate qualitative aspects of the selected feature sets, as shown in Fig. \ref{fig:stabilities}. The demonstrated stabilities and redundancy rates (RED) of the feature sets selected by RENT and UBayFS across the five folds tend to increase with $\max_s$. While RENT has a slightly lower and more fluctuating stability (around 0.5), UBayFS shows a clear convergence at around 0.6. The RED is below $0.25$ for all possible numbers of features, indicating that both RENT and UBayFS select features with small correlations.

\subsection{Experiment 2: feature selection with prior knowledge}

Previous research on GEP NEN shows that some features impact the survival of patients; those are \textit{Age at diagnosis, WHO performance status, Primary tumor location, Tumor morphology, Tumor differentiation, Lactate dehydrogenase (LDH), Platelets, Albumin, Ki-67, SUV$_{\max}$}, and \textit{TNM-staging} ~\cite{henning05,stokmo2022gastro,henning09,henning10,henning11,henning12,henning13,henning14}. Tumor differentiation is highly correlated to tumor morphology, so we do not include the feature in this work. Furthermore, findings by \cite{stokmo2022gastro} indicate a high relevance of the features \textit{Total MTV [cm$^3$] and Total TLG [g]}, which shall be investigated. 

In this experiment, we focus on these features (a total number of 22 features in the encoded space) within our feature selection and prediction pipeline. In particular, during experiment 1, the aforementioned features comprise 30\% of the final feature sets (on average across the five folds and given $\max_s=20$ features, each). We refer to this score as \text{PERC} (percentage of selected features supported by literature). In the following, we deploy prior weights on these features to investigate how UBayFS as a hybrid feature selector combining information from experts and data, performs in comparison to the pure data-driven methods presented in experiment 1. Since RENT cannot incorporate prior feature importances, this evaluation is restricted to UBayFS. 

Specifically, we increase the prior weight of the 22 features supported by literature (referred to as prior-elevated features) to the following levels: $w\in \{0.1, 10, 20, \dots, 100, 110\}$ --- after evaluating all levels with respect to predictive performance, we restrict to special cases $w=0.1$ (non-informative prior weighting), $w=50$ (mediocre prior weighting), and $w=110$ (strong prior weighting). After applying UBayFS with the given levels of prior information, we examine how the feature set and the predictive performance develop. The case of $0.1$ is equivalent to the uniform case without prior knowledge (default setup for UBayFS in experiment 1). In contrast, prior weight 110 indicates that each prior-elevated feature already is assigned a higher score than the maximum score that can be achieved throughout the elementary models ($M=100$) --- as a result, the selected features are exclusively restricted to those with prior information and elementary feature selectors in UBayFS are only used to select a feature set of $\max_s=20$ features among the 22 prior-elevated features. 

\begin{figure}
    \centering
    \begin{subfigure}{0.45\textwidth}
        \centering
        \includegraphics[width=\textwidth]{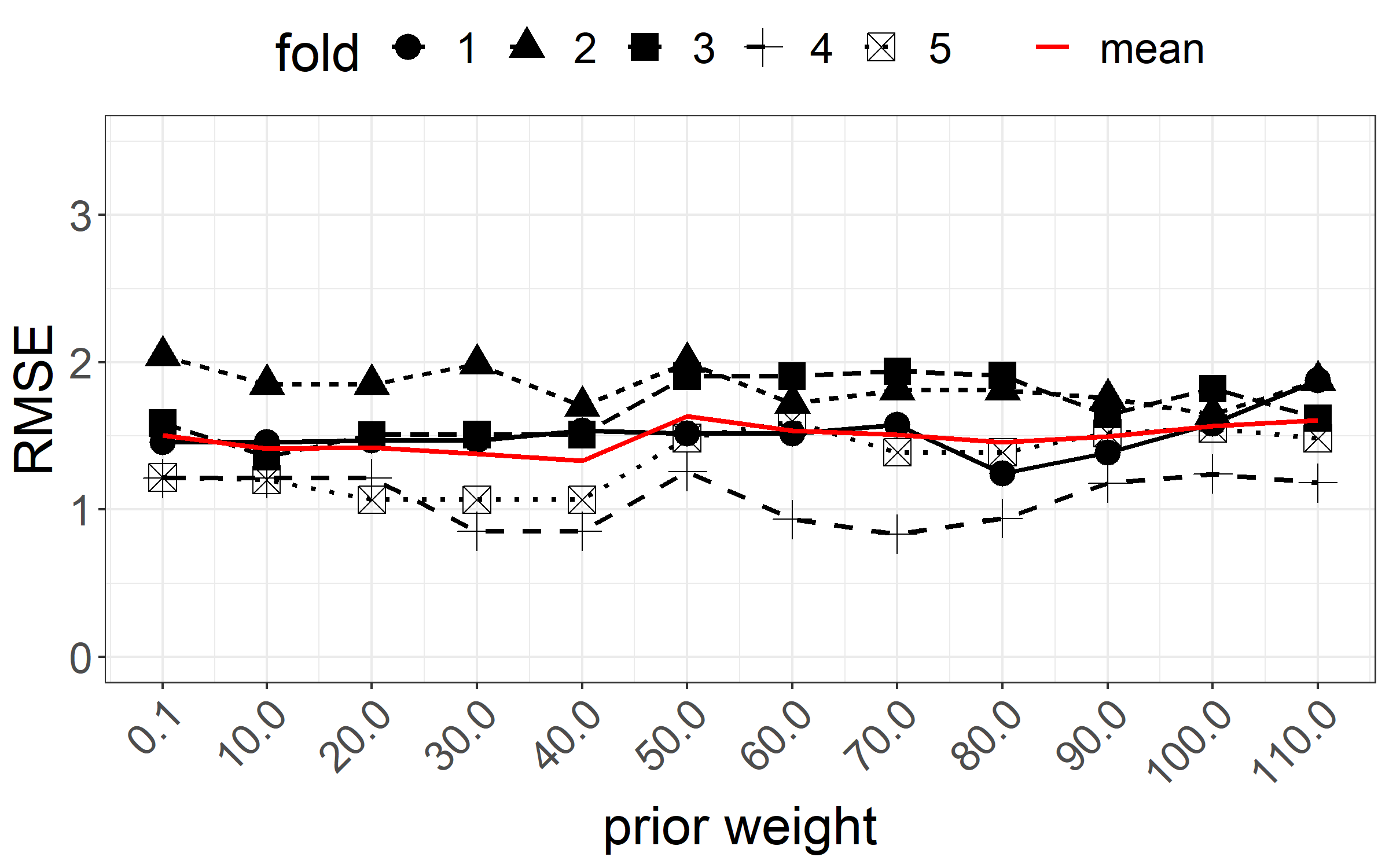}
        \caption{Predictive performance using $k$NN regression}
    \end{subfigure}
    \begin{subfigure}{0.45\textwidth}
        \centering
        \includegraphics[width=\textwidth]{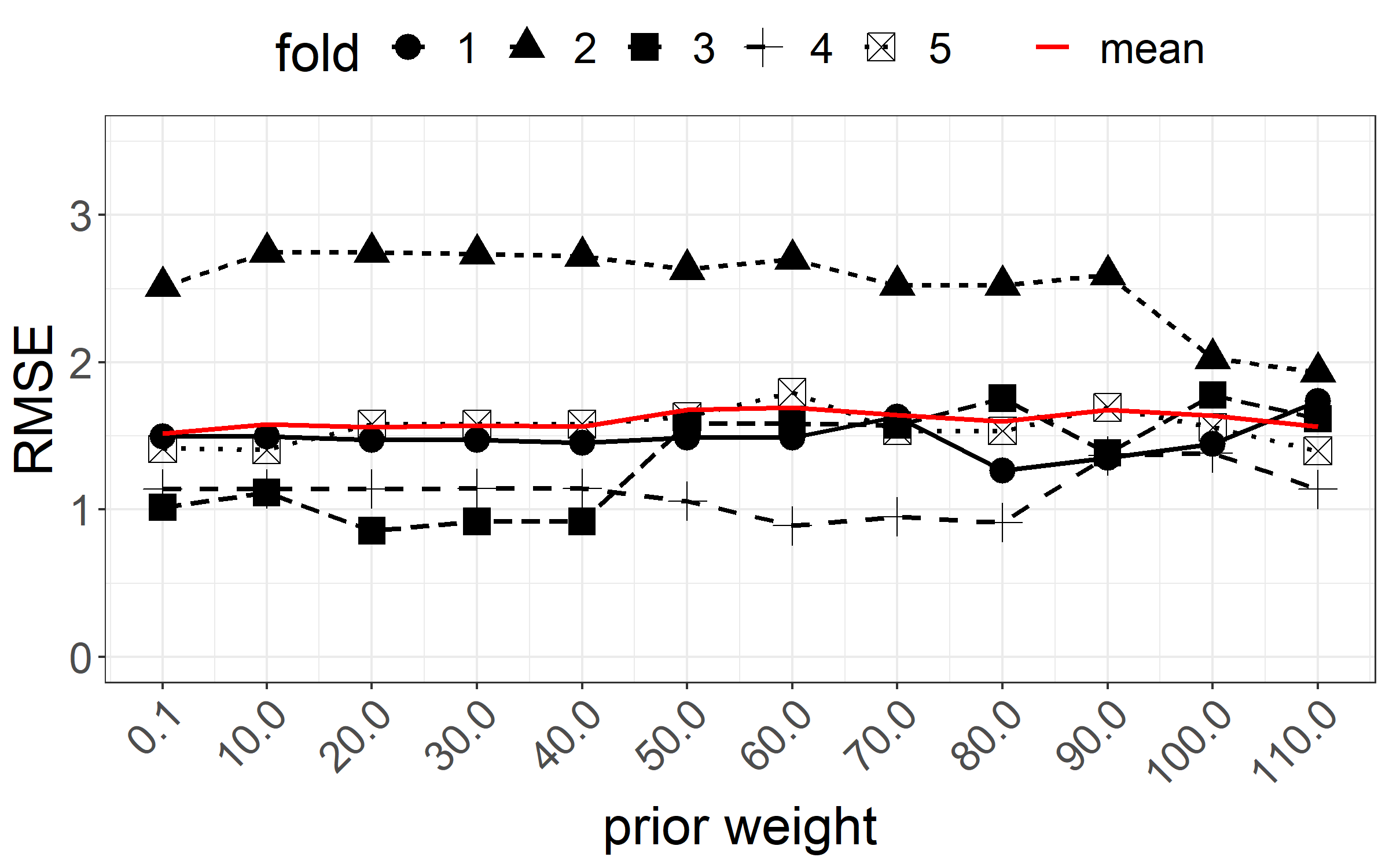}
        \caption{Predictive performance using linear regression}
    \end{subfigure}
    \begin{subfigure}{0.45\textwidth}
        \centering
        \includegraphics[width=\textwidth]{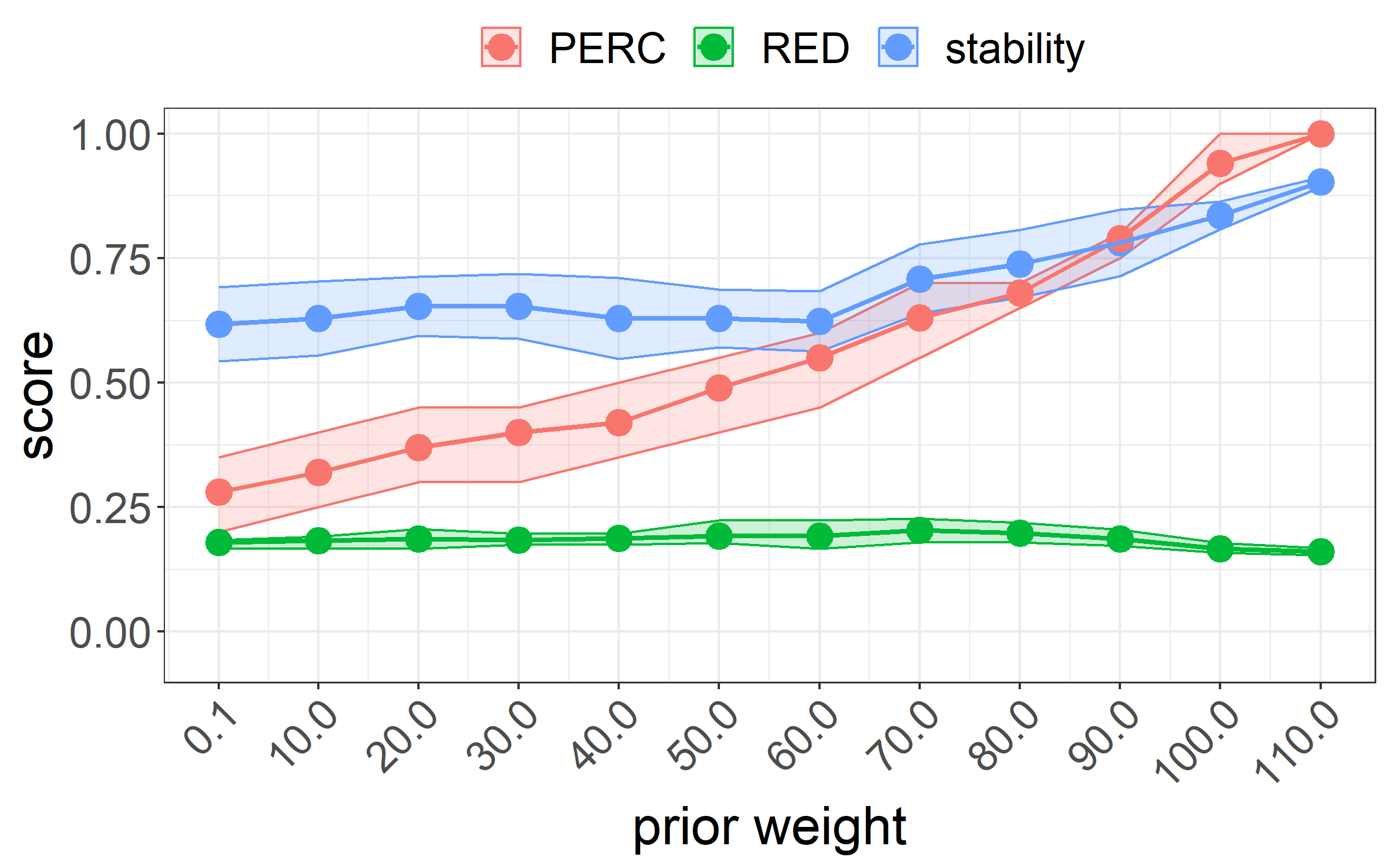}
        \caption{Qualitative evaluation}
    \end{subfigure}
    \caption{Experiment 2: predictive performances on fold 1-5, and qualitative metrics of features sets produced by UBayFS at different levels of prior knowledge on features with evidence from literature ($\max_s=20$).}
    \label{fig:performance_exp2}
\end{figure}

\paragraph{Predictive performance} Fig. \ref{fig:performance_exp2} shows the average performances along with the standard deviations across the 5 test folds. In general, lower levels of prior weights do not significantly impact the performance, although a minor improvement can be observed in folds 4 ($k$NN) and fold 3 (linear model) up to $w=40$. By increasing the prior weight to a higher level, performance levels lead to stronger variability and an increase of RMSE in the better-performing folds, such as fold 4. Finally, if the prior weight is set to the maximum level of $110$, all folds converge to a similar level since the data-driven feature selection hardly contributes to these setups. Thus, a potential conclusion is that moderate levels of prior knowledge can slightly increase models' capabilities. In contrast, strong prior knowledge leads to a convergence towards the global mean performance across all folds --- such prior setup acts as a strong restriction of the search space exploited by the feature selector.

\paragraph{Stability} In contrast to the minor effects of prior knowledge on predictive performance, stability increases significantly, as shown in Fig. \ref{fig:performance_exp2}. Finally, at a maximum level of $w=110$, stability converges towards an almost perfectly stable solution. This is due to the restriction of the search space to the prior-elevated features, which results in a selection of 20 out of only 22 features in total. As expected, the percentage of selected features supported by literature (PERC) also increases linearly with the level of prior weights provided. The redundancy rate between the selected features shows a slight decrease, indicating that the prior-elevated features contain only small correlations.

\section{Discussion}

\paragraph{Experiment 1} In our first experiment, we left out prior expert knowledge and let the feature selection be purely data-driven. We know that certain features were prognostic for survival in earlier studies, as mentioned in experiment 2 below. We wanted to study whether the same prognostic features would still be selected and if there were any currently unknown prognostic features that could be further researched. Comparing the two first columns in Tab. \ref{tab:feature_ranking_exp1} we can see which features are selected repeatedly in different folds with RENT and UBayFS. We must keep in mind that we cannot directly compare the importance of the features in terms of a coefficient (e.g similar to Cox regression), just that they are repeatedly selected in each fold. Further, the correlation between features must also be considered when comparing the importance of features which we can find in Fig. \ref{fig:correlations}. Two or more features with a moderate/high correlation contain the same information with respect to the model, and one fold may choose one over the other, whilst another fold may choose a highly correlated one instead. This results in a lower number for both features, not reflecting the importance of both when comparing them with a feature with a high number.

In block (p) (baseline patient characteristics) we have a few features that one would expect to be prognostic for OS. One obvious one would be \textit{TNM-stage IV} disease which does not seem to be chosen at all by RENT and UBayFS. But looking at the correlation heatmap in Fig. \ref{fig:correlations} we see that this feature is highly correlated to several other features, among those \textit{Metastatic Disease at Time of Diagnosis} and \textit{Treatment Intention Palliative}. We see that this last one gets chosen four out of five times with RENT and five out of five times with UBayFS which probably explains why \textit{TNM-stage IV} does not seem to be important. Having a palliative treatment intention usually means you have stage IV disease. This is also a well-known prognostic indicator from the literature \cite{henning06}. Bone metastasis is usually a poor prognostic indicator in several types of cancers \cite{henning28} and it is not surprising that this is chosen all the time. We also know that \textit{WHO PS} is a prognostic indicator in these patients. This is also reflected in the number of folds it is chosen by RENT and UBayFS, but it is only WHO level 2 that seems to be important. That said, Fig. \ref{fig:correlations} shows that WHO levels 3 and 4 are highly correlated to some of the SUV-parameters which might contribute to those never being selected. \textit{Radical Surgery} is quite often chosen by both RENT and UBayFS and is also a predictable prognostic indicator. Having radical surgery means that all viable tumors are removed, and that is only possible if you have a low tumor burden. This underlines the importance of surgery in the curative intended treatment of this type of cancer.

Next, in block (b) (baseline blood values), we see that both \textit{CRP} and \textit{ALP $>$ Normal $<$= 3UNL} get selected equally many times by both RENT and UBayFS, and both have a high number indicating importance over the other features in this block. A high CRP at baseline has previously been shown to be a poor prognostic feature in some studies \cite{henning05,henning19,henning20}, whilst others have not replicated this \cite{henning21}. This is probably not surprising as this has been shown to be a poor prognostic indicator in advanced cancer patients in a palliative setting, and especially in GEP NEN \cite{henning22,henning23,henning24,henning25}. \textit{ALP} has also been shown in studies to be prognostic for a shorter OS \cite{henning05,henning26,henning27}. For \textit{Albumin} and \textit{Platelets}, RENT chose these only half as many times as UBayFS. Both have been shown to be prognostic indicators of OS \cite{henning05,henning06}. Interestingly, \textit{Haemoglobin}, \textit{WBC}, \textit{LDH}, and \textit{Chromogranin A} are barely chosen or are not chosen neither by RENT nor UBayFS. All these features have previously been shown to be prognostic for OS \cite{henning05}.

Moving on to block (h) (re-evaluated histology) we have quite a few features that are well-known prognostic indicators. The strongest one from the literature is probably \textit{Ki-67} which is used in the classification system of NEN. The second strongest is probably \textit{Tumor Morphology} which has been shown in several studies to be prognostic for OS \cite{henning05,henning06}. We see that \textit{Ki-67} is chosen every time from all five folds both for RENT and UBayFS supporting this feature as a strong prognostic indicator for OS. Further, \textit{Tumor Morphology} gets chosen four out of five times with RENT and three out of five times with UBayFS. This is also to be expected since we know that patients with NET G3 have a better OS than those patients with NEC \cite{henning29}. What is surprising is that most tumor sites, especially those patients with unknown primary and esophagus NEN, are not chosen by RENT or UBayFS. \textit{Primary Tumor Site} has been shown to be prognostic in several studies \cite{henning05,henning06}. Several of the features like \textit{Stroma, Architecture, Vessel Pattern, Co-existing neoplasm}, and \textit{Geographic Necrosis} are considered typical for either NET G3 or NEC \cite{henning30}, and one might assume these are highly correlated with \textit{Tumor Morphology}. Although this is not reflected in Fig. \ref{fig:correlations}. Almost none of these features are chosen with RENT or UBayFS except for \textit{Stroma}. NET G3 typically have hyalinized stroma and NEC have desmoplastic stroma \cite{henning30}.

Further, in block (i) (PET/CT imaging) the interesting features are \textit{Total  MTV}, \textit{Total TLG}, and the SUV-parameters. From Fig. \ref{fig:correlations} and previous literature \cite{stokmo2022gastro} we know that these features are often (if not always) highly correlated. Hence, the selection of \textit{SUV$_{\max}$ (total)} instead of the other features is probably related to this. Moreover, we know from previous studies \cite{stokmo2022gastro, henning11,henning12,henning13} that global measures such as \textit{Total MTV} and \textit{Total TLG} are poor prognostic indicators for OS in these tumors, but we lack stronger evidence in form of larger studies. Here we see that \textit{SUV$_{\max}$ (total)} is chosen in all five folds both for RENT and UBayFS supporting the previous findings that PET-parameters are good prognostic features of OS.

Finally, in block (t) (treatment) we can see a few features are selected often. \textit{Chemotherapy treatment with cisplatin/etoposide} is not surprisingly a predictor for OS, and most of the patients did indeed receive this combination. No chemotherapy is obviously detrimental. We also see that the \textit{Chemotherapy treatment with temozolomide/everolimus} gets chosen often both by RENT and UBayFS. This is probably because this chemotherapy regimen is more often chosen for those patients with a low \textit{Ki-67} and these are more likely to be NET G3 which already have a better OS. Further, both \textit{Number of Courses} and \textit{Progression} are two features that are selected often by RENT and UBayFS. \textit{Progression} and \textit{No Progression} are obviously poor prognostic indicators, and one could assume that the higher \textit{Number of Courses} a patient receives the longer before they have progression and hence they live longer. This is of course only an assumption and interpretation of the data at hand. It is a bit surprising that the response evaluation results did not get chosen. One would assume that patients with the best response - stable disease would fare better than those with progressive disease. Looking at Fig. \ref{fig:correlations} the features from this block have low correlation coefficients.

\paragraph{Experiment 2} Here we added prior expert knowledge and assigned two different weights. A weight $w$ = 50 means approximately 50\% expert-driven and 50\% data-driven. A weight $w$ = 110 means almost purely an expert-driven approach where we effectively force the selection of features only from the subset of those from prior expert knowledge. We concentrated on features that are well documented in several previous studies, although there exist more features in the literature suggesting prognostic values than these. The features selected from prior expert knowledge are listed in the first paragraph in Section 3.4 and marked by an asterisk in Tab. \ref{tab:feature_ranking_exp1}.

If we concentrate on the second, third, and fourth columns, which shows the difference between roughly 0\%, 50\% and 100\% expert-driven, we see that none of the marked features drops in importance as we increase the value of expert knowledge. Some features that were never chosen with a pure data-driven model are still not chosen. One could argue that these are probably not strong features to begin with, or that other features contain the same and/or stronger information. A few features only get chosen when almost completely removing the data-driven part and make a huge leap from not being chosen to being chosen five times. We argue that one should be careful to put too much importance on these features as we expect these are more or less forced to be chosen. 

A few features stand out by being stable across all values of $w$; \textit{WHO Performance Status}, \textit{Albumin}, \textit{Platelets}, \textit{Ki-67}, \textit{Tumor Morphology}, \textit{Total MTV}, \textit{Total TLG}, and \textit{SUV$_{\max}$}. It would be bold to assume that these features are the most important and stable predictors of OS from the subset of expert knowledge markers, but that would probably be too premature. Further, it is also interesting to notice that even though several parameters from PET are highly correlated, several are still chosen very often by the model. This is in line with the results of our previous study (\cite{stokmo2022gastro}. Moreover, it is a bit surprising that \textit{Primary Tumor Site}, especially \textit{Unknown Primary} and \textit{Esophagus}, is not chosen more often as these are well-known negative predictors of OS \cite{henning05,henning06}.

We also notice that some of the other non-marked features drop in importance as we increase $w$, and this is probably related to the fact that the features overlap in the information they add to the model. A few of these features are also moderately or highly correlated. E.g. \textit{CRP} is correlated with quite a few of the other blood markers, and this could explain why it falls in importance when increasing $w$. \textit{Mets Bone} (bone metastases) is not listed in the correlation heatmap and thus has no moderate or high correlations with other features, but still completely falls out. Bone metastases usually occur late in several cancers and is a poor prognostic feature. Hence, one should assume that this feature and similar ones like \textit{CRP}, \textit{ALP}, which performs well with low values of $w$ falls of in the pure knowledge-driven model because the model is "forced" to select only marked features. We must remember that the $w$ = 110 is an extreme expert-driven model which is probably not clinically relevant but was added to explore and evaluate what the model did in this extreme situation. This is a small, novel study with few patients and really the first of its kind for exploring and evaluating RENT and UBayFS on clinical data. Using these ensemble feature selectors may be used for validating already established features, or to find new features not previously known. Evaluation into which $w$ is optimal should be explored further in future studies.

\section{Conclusion}

In conclusion, although we cannot ascertain how important different features are compared to each other and if they contribute to poorer or better survival, we do find similar results as several previous studies. The most stable and predictive features in our study are \textit{WHO Performance Status}, \textit{Albumin}, \textit{Platelets}, \textit{Ki-67}, \textit{Tumor Morphology}, \textit{Total MTV}, \textit{Total TLG}, and \textit{SUV$_{\max}$}. 

From a data science perspective, we demonstrated the capabilities of the ensemble feature selection techniques RENT and UBayFS for healthcare problems --- in particular, the inclusion and comparison of expert- and data-driven setups, as well as combinations of both, allow the user to gain relevant information for clinical use.


\bibliographystyle{unsrt}
\bibliography{literature}

\end{document}